\documentclass{article}

\PassOptionsToPackage{numbers, compress}{natbib}
\usepackage[preprint]{neurips_2026}
\usepackage[utf8]{inputenc}
\usepackage[T1]{fontenc}
\usepackage{adjustbox}
\usepackage[normalem]{ulem}   % for \sout (strikethrough)
\usepackage{xcolor}           % colour support
\usepackage{microtype}
\usepackage{graphicx}
\usepackage{subcaption}
\usepackage{booktabs}

\usepackage{tcolorbox}

\usepackage{hyperref}
\hypersetup{
    colorlinks=true,
    linkcolor=blue,
    filecolor=magenta,
    urlcolor=cyan,
    pdftitle={Overleaf Example},
    pdfpagemode=FullScreen,
}
\usepackage{url}
\usepackage{mathrsfs}
\usepackage{amsmath}
\usepackage{amssymb}
\usepackage{mathtools}
\usepackage{amsthm}

\usepackage[capitalize,noabbrev]{cleveref}

\usepackage{tikz}
\usetikzlibrary{positioning,calc,arrows.meta}

\usepackage[textsize=tiny]{todonotes}

\usepackage{amsmath,amsfonts,bm}

\def\eqref#1{equation~\ref{#1}}
\def\1{\bm{1}}

\DeclareMathAlphabet{\mathsfit}{\encodingdefault}{\sfdefault}{m}{sl}
\SetMathAlphabet{\mathsfit}{bold}{\encodingdefault}{\sfdefault}{bx}{n}

\theoremstyle{plain}

\theoremstyle{definition}

\theoremstyle{remark}

 \title{Sampling Data with Chains of Forward-Backward Diffusion Steps}

\author{%
Hyunmo Kang \thanks{Co-first authors. Contact: hkang56@jh.edu, mwyart1@jh.edu}\\
Johns Hopkins University
\And
Noam Itzhak Levi$^{*}$\\
EPFL
\And
Corinna Elena Wegner$^{*}$\\
EPFL \& University of Göttingen
\AND
Daniel J. Korchinski$^{*}$\\
EPFL
\And
Matthieu Wyart\\
Johns Hopkins University \& EPFL
}

\begin{document}
\maketitle

%%%%%%%%%%%%%%%%%%%%%%%%%%%%%%%%%%%%%%%%%%%%%%%%%%%%%%%%%%%%
%                       ABSTRACT
%%%%%%%%%%%%%%%%%%%%%%%%%%%%%%%%%%%%%%%%%%%%%%%%%%%%%%%%%%%%

\begin{abstract}

Sampling from learned high-dimensional distributions is a foundational
computational problem. We introduce \emph{U-turn chains}: Markov chains
obtained by iterating short forward--backward steps of a diffusion model,
in which each step proposes a move that remains on the learned data
manifold and, paired with a Metropolis--Hastings correction, samples from
energy-modified targets. For synthetic languages, we show that minimal U-turn
dynamics undergoes an ergodicity-breaking phase transition driven by
fragmentation of the data manifold; ergodicity is restored at larger
U-turn magnitude. In the non-ergodic regime, low-level features relax
faster than high-level ones, an ordering that inverts only at
sufficiently large U-turn magnitude. We test these predictions on
natural language 
and natural images. In both
modalities, minimal U-turns relax slowly, especially for high-level
features approximated by deep representations in CNNs or LLMs. The layer-ordering inversion appears only at large
noise when mixing is efficient- signatures consistent with strongly constrained, weakly mixing
local dynamics. We  discuss the implications of these results for sampling with diffusion models.

%We introduce U-turn chain dynamics, obtained by iterating short forward--backward steps with a diffusion model.
%Starting from a data sample, each step partially corrupts the sample and then reconstructs it, defining a Markov chain on the learned data distribution.
%We use this dynamics to probe the geometry and hierarchical organization of data. In a synthetic Random Hierarchy Model, we show that minimal U-turn dynamics can undergo an ergodicity-breaking transition: for sufficiently structured data, local moves explore only a restricted component of the data manifold, while increasing the U-turn magnitude restores ergodicity.
%The model also predicts that higher-level latent variables relax more slowly than lower-level ones in the small-step regime, and that this layer-wise ordering can invert only at sufficiently large U-turn magnitude.
%We test these predictions in natural language using a masked diffusion language model and LLM representations.
%For minimal U-turn steps, deeper language representations relax much more slowly than earlier ones.
%Moreover, both inversion of layer-wise relaxation and the peak of the dynamical susceptibility occur only at large masking fraction $\rho$.
%By analogy with the Random Hierarchy Model, these signatures are consistent with a highly structured regime in which minimal U-turn dynamics is strongly constrained and may be effectively non-ergodic or very slowly mixing.
%As a complementary image experiment, we show that CNN representations along image U-turn chains display analogous depth-ordered relaxation across noise levels.
\end{abstract}

\section{Introduction}

% \begin{itemize}

%     \item latent variable models

%     \item h

% \end{itemize}

\begin{figure}[htbp]
    \centering
    \begin{subfigure}{0.28\linewidth}
        \centering
        \includegraphics[width=\linewidth]{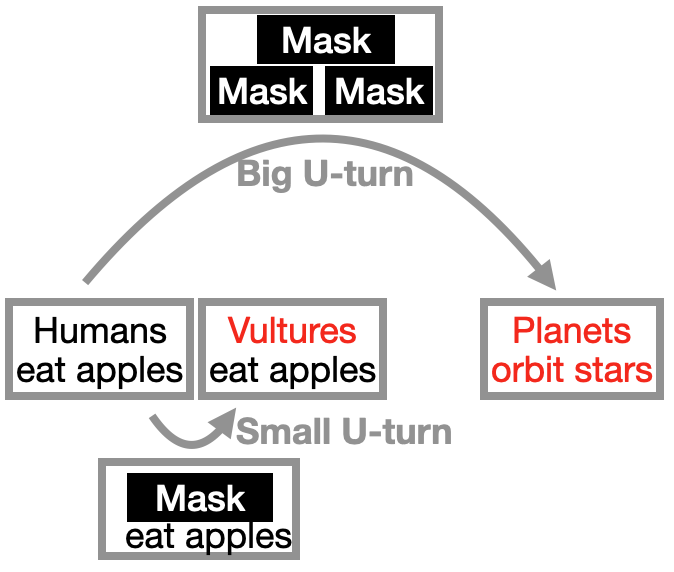}
        % \caption{Unsteered sampling}
        \label{fig:mc_unsteer}
    \end{subfigure}
    % \hfill
    \begin{subfigure}{0.33\linewidth}
        \centering
        % \vspace{1cm}
        \includegraphics[width=\linewidth]{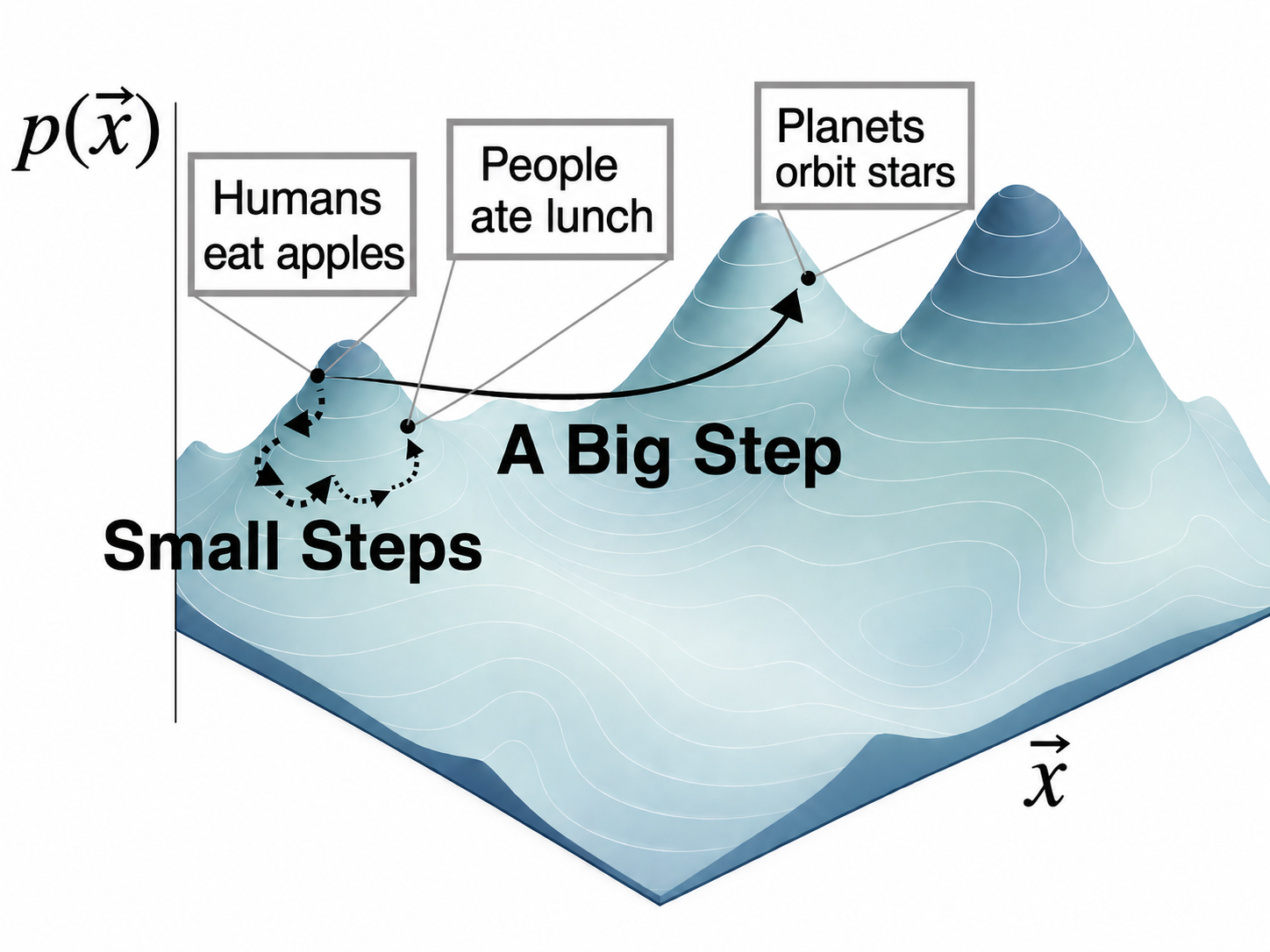}
        \label{fig:mc_ergodic_hierarchy}
    \end{subfigure}
    \hspace{0.2cm}
    \begin{subfigure}{0.26\linewidth}
        \centering
        \includegraphics[width=\linewidth]{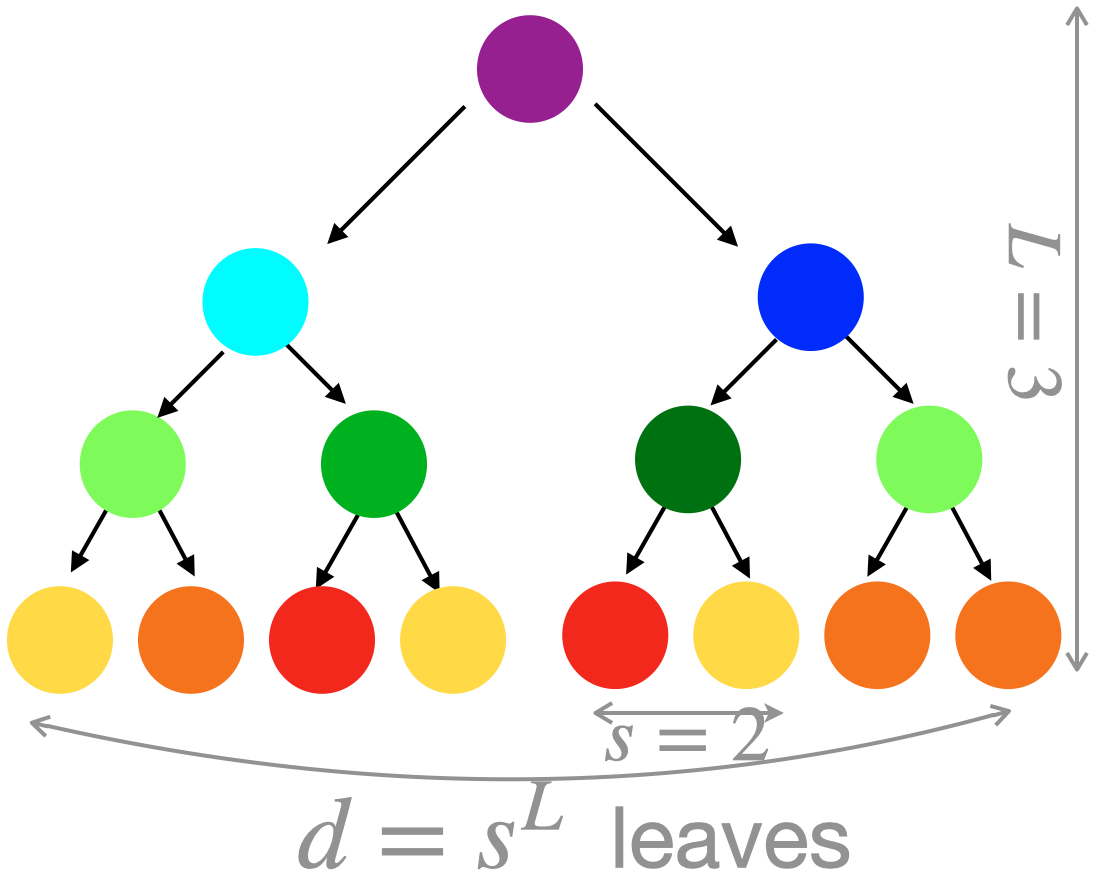}
        % \caption{Steered sampling}
        \label{fig:mc_steer_dog_cat}
    \end{subfigure}
\caption{\textbf{Left:} A single U-turn move first corrupts a sample by adding noise or masking part of the input, then reconstructs it using a trained diffusion model.
The U-turn magnitude controls the size of the perturbation.
\textbf{Middle:} Iterating U-turn moves defines a Markov chain on the learned data distribution, allowing us to study whether the chain is ergodic.
\textbf{Right:} Schematic of the Random Hierarchy Model, where visible tokens are generated from a hierarchy of latent variables through recursive production rules.
This synthetic model provides a controlled setting for analyzing ergodicity and layer-wise relaxation.
}
    \label{fig:mc_sampling}
\end{figure}

Sampling from a high-dimensional probability distribution is a foundational
computational problem across the sciences, underlying Bayesian inference
\cite{robert2004monte}, statistical physics
\cite{newman1999monte,frenkel2002understanding}, and molecular simulation \cite{BERNARDI2015872, Oruchic1997}.
The difficulty is twofold. First, the volume of the state space grows
exponentially with its dimension, so uniform exploration is hopeless without
exploiting structure. Second, distributions of practical interest are
typically multimodal: high-probability regions are separated by low-density
barriers that trap local-move dynamics for times that grow rapidly with the
barrier height and the dimension
\cite{mezard1987spin,levin2009markov}.
Classical Markov chain Monte Carlo (MCMC) algorithms - Metropolis-Hastings
\cite{metropolis1953equation,hastings1970monte}, Langevin dynamics
\cite{welling2011bayesian}, Hamiltonian Monte Carlo \cite{neal2011mcmc},
and replica exchange \cite{swendsen1986replica,earl2005parallel}- partially
mitigate this slowdown, but their efficiency degrades sharply when the
target distribution is concentrated on an unknown low-dimensional manifold,
as is the case for natural images and text.

Diffusion models \cite{sohl2015deep,ho2020denoising,song2021score}
implicitly learn such a manifold from data, suggesting a complementary
route to sampling: rather than designing proposals in the ambient space,
one can use the diffusion model itself to propose moves that remain on the
learned support. Combined with a Metropolis--Hastings correction, this
yields a sampler from energy-modified distributions of the form
$P_H(x) \propto P(x)\,e^{-H(x)}$, where $H(x)$ is a user-specified bias
encoding desired attributes.

Diffusion models are typically used to generate samples from pure noise,
producing global moves that preclude local exploration around a specific
region of interest - and, in particular, around the minima of $H(x)$.
They can however also be used to perform forward--backward transformations,
or U-turns~\cite{nichol2021improved,dhariwal2021diffusion,behjoo2023u,sclocchi2024phase},
in which a datum is stochastically perturbed by first adding a controllable
amount of noise and then denoising it. In this work, we introduce
\emph{U-turn chains}, obtained by iterating such forward--backward
transformations: starting from a sample $x$, a forward diffusion step
produces a partially corrupted representation $x_t$, and a reverse
denoising step yields a new sample $x'$. Repeating this procedure defines
a Markov chain on the learned data distribution.

As a first step, this work focuses on the fundamental questions of ergodicity and
relaxation in the absence of steering ($H=0$): under what conditions do
U-turn chains explore the data manifold ergodically, and how fast do
different levels of structure in the data relax along the chain?
Many data domains - including language and images - exhibit hierarchical
compositional structure that can be approximated by tree-like generative
models~\cite{chomsky2014aspects,jager2012formal,grenander1996elements},
a structure that has been argued to be central for learning generative
models themselves~\cite{mossel2016deep,poggio2017why,malach2018provably,%
schmidt2020nonparametric,malach2020implications,cagnetta2024deep,%
cagnetta2024towards,cagnettaDerivingNeuralScaling2026}.
Such data admit multiple levels of description, from low-level features
(pixels, edges) to high-level abstractions (captions, semantic content),
raising the question of how mixing times depend on the level considered.

To make these questions tractable theoretically, we work with the Random
Hierarchy Model (RHM)~\cite{cagnetta2024deep}, a simplified probabilistic
context-free grammar~\cite{grenander1996elements} with
fixed tree topology and random production rules. Despite its simplicity,
the RHM has produced quantitatively predictive theories of neural scaling
laws~\cite{cagnetta2024towards,
cagnettaDerivingNeuralScaling2026, cagnetta2025learningcurvestheoryhierarchically} and of compositional generalization in
diffusion models for both language and
images~\cite{sclocchi2024phase,sclocchi2024probing}.
In this model, we show that (i) U-turn chains undergo a phase transition
at which ergodicity is broken, and (ii) in the non-ergodic regime
low-level features relax faster than high-level ones---an ordering that
inverts deep in the ergodic phase. We then test these predictions on
natural language and on images: in both domains, mixing under minimal
U-turns is slow, particularly for high-level latents, consistent with a
non-ergodic or weakly ergodic local dynamics whose ergodicity is restored
by increasing the U-turn magnitude.

\subsection{Additional related works}

\textbf{Characterizing data structure via diffusion models.}
\cite{guth2026learningnormalizedimagedensities} introduce a diffusion-inspired dual score-matching framework that enables the computation of normalized image densities, providing direct access to the energy landscape of natural images. Closer to our work, \cite{sclocchi2024probing,sclocchi2024phase} show that a single U-turn can be used to reveal hierarchical structure in data. These works also demonstrate that the representation of CNNs at different depths can be used as a proxy of features or latent variables at different levels of abstraction- a view we confirm here and extend to LLM representations in the language domain.
U-turn chains generalize this approach, allowing not only the study of such structure including connectedness and ergodicity, but also the steering of samples.

\textbf{Monte Carlo approaches with diffusion models.}
A Metropolis–Hastings framework based on diffusion models is introduced in \cite{huntsmith2023acceleratingmarkovchainmonte}, where proposals are generated by denoising from pure Gaussian noise. Such global proposals are ill-suited for studying the local properties of the data distribution considered here, or for identifying local optima of a generic bias $H(x)$, which requires local exploration.

% \textbf{Steering.}
% Steering toward desired features is typically achieved by modifying the reverse diffusion process \cite{li2022diffusionlm,austin2021structured,dhariwal2021diffusion,ho2022classifierfree}. However, these methods do not allow for continuous refinement of a given datum; instead, they generate new samples. More closely related to our approach, SDEdit \cite{meng2022sdedit} performs image manipulation by applying a single forward–backward diffusion step to an input image, but is limited to one such step. Closest to our work, Mix-and-Match \cite{mireshghallah2022mixmatchlearningfreecontrollable} formulates controllable text generation as sampling from an energy-based language model using Metropolis–Hastings with BERT-based proposals, and \cite{forristal2023blockmetropolishastingssamplercontrollable} extend this idea using larger block proposals generated by large language models.
% Our work extends this line of research by leveraging diffusion models to study both text and images, providing a theoretical framework.

\section{General Formalism}
\label{sec:formalism}

\subsection{ U-turns or forward-backward process}

Let $x \sim P(x)$ denote a data sample. Diffusion models define a corruption process mapping a clean sample $x$ to a partially corrupted representation $x_t$, where $t$ denotes the noise level. The interpretation of $t$ depends on the diffusion model. For discrete diffusion, $t$ corresponds to the number of masked tokens. For continuous diffusion, $t \in [0,T]$ corresponds to the time of the forward diffusion process. 

The forward process defines a conditional distribution $q(x_t \mid x)$, and the trained model approximates the reverse conditional
$p_\theta(x \mid x_t)$. We define a \emph{U-turn move} \cite{nichol2021improved,dhariwal2021diffusion,behjoo2023u,sclocchi2024phase} of size $t$ as $x \rightarrow x_t \rightarrow x'$,
with $x_t \sim q(x_t \mid x)$,  $x' \sim p_\theta(x \mid x_t)$.
This induces the transition kernel
\[
U_t(x',x)
=
\int q(x_t \mid x)\, p_\theta(x' \mid x_t)\, dx_t .
\]

The parameter $t$ controls the scale of the move: small $t$ produces local modifications, while larger $t$ allows more global changes. Crucially, regardless of step size, U-turn proposals remain on the data manifold learned by the diffusion model.
% [DELETED author comment - hyunmo: Maybe we can rephrase this...]
The kernel $U_t(x',x)$ naturally defines a Monte Carlo sequence obtained by
iterating U-turn moves. Starting from an initial sample $x^{(0)}$, we generate a \textbf{U-turn chain} $x^{(0)} \rightarrow x^{(1)} \rightarrow x^{(2)} \rightarrow \cdots
$ where each step is obtained by
$x^{(n+1)} \sim U_t(\cdot \mid x^{(n)})$.

%Equivalently, each iteration consists of sampling
%\[
%x_t \sim q(x_t \mid x^{(n)}), \qquad
%x^{(n+1)} \sim p_\theta(x \mid x_t).
%\]

%This defines a Markov chain on the space of samples. The next section
%characterizes its stationary distribution.
\subsection{Stationary distribution of the UTMC}

\paragraph{Proposition 1.}
Assume the diffusion model is exact, i.e.\ $p_\theta(x \mid x_t)=p(x \mid x_t)$. Then the U-turn kernel satisfies detailed balance with respect to $P(x)$:
$
U_t(x',x)P(x)=U_t(x,x')P(x').$
If the resulting Markov chain is ergodic, its stationary distribution is $P(x)$. 

\paragraph{Proof (sketch).} 
Using Bayes' rule $p(x'|x_t)=\frac{q(x_t|x')P(x')}{p(x_t)}$
in the definition of $U_t$ yields
\[
U_t(x',x)P(x)
=
\int
\frac{q(x_t|x)q(x_t|x')}
     {p(x_t)}
P(x)P(x')\,dx_t,
\]
which is symmetric in $x$ and $x'$. $\square$

\subsection{Energy-modified sampling}

Although the main focus of this paper is the intrinsic dynamics of UTMC, the detailed-balance property also makes them useful as proposal kernels for sampling from energy-modified distributions. Suppose we wish to sample from the modified distribution $P_H(x) \propto P(x)e^{-H(x)}$,
where $H(x)$ is an additional energy. Using U-turn proposals within a Metropolis--Hastings scheme, a proposal
$x' \sim U_t(\cdot|x)$ is accepted with probability
\[
A(x \to x')
=
\min\!\left(
1,
\frac{P_H(x')U_t(x,x')}
     {P_H(x)U_t(x',x)}
\right)=
\min\!\left(1, e^{-(H(x')-H(x))}\right).
\]
Under ergodicity, this Markov chain converges to $P_H(x)$. We study below under which conditions ergodicity may be reached.

\section{Ergodicity and mixing in the Random Hierarchy Model}
\label{sec:theory}
\subsection{Synthetic hierarchical data: the Random Hierarchy Model}

The \textbf{Random Hierarchy Model (RHM)} is a synthetic language, belonging to the class of probabilistic context-free grammars \cite{rozenberg_handbook_1997, cagnetta2024deep}. Data are produced on a rooted regular tree of depth $L$ and branching factor $s$ as shown in Fig.~\ref{fig:mc_sampling} Right. Levels are indexed from the visible layer upward:
$\ell=0$ denotes the leaves and $\ell=L$ the root. At level $\ell$ there are
$s^{L-\ell}$ variables $x_\ell^i$, each taking values in a vocabulary of size
$v$. The visible sequence is $x_0=(x_0^1,\dots,x_0^d)$ with $d:=s^L$.

Each parent symbol at level $\ell+1$ is associated to $m$ production rules, each
mapping the parent to an $s$-tuple of child symbols at level $\ell$. Rules are
sampled uniformly without replacement and kept fixed for a given RHM instance.
Generation starts from a uniformly sampled root symbol; each internal node then
chooses one of its $m$ rules uniformly to generate its children. We impose the
unambiguity condition that no two distinct parent symbols generate the
same child $s$-tuple, so every valid visible sentence has a unique latent tree.
All valid sentences are therefore equiprobable.

The rule density
\[
f:=\frac{vm}{v^s}=\frac{m}{v^{s-1}}
\]
is the fraction of possible child $s$-tuples used by the grammar. When $f=1$,
all tuples are allowed and the hierarchy imposes no constraint on visible
strings. Smaller $f$ makes the grammar sparser and the data distribution more
structured. Given a partially masked visible sequence, the posterior over
both visible and latent variables can be computed exactly by Belief Propagation (BP) \cite{mezard2009information, sclocchi2024phase}. We use BP as a
Bayes-optimal denoiser, so the RHM dynamics reflect perfectly the data
distribution rather than the neural approximation error of a trained diffusion model.

\subsection{Observables at different hierarchical levels}

\begin{figure}[htbp]
\centering
\includegraphics[width=\linewidth]{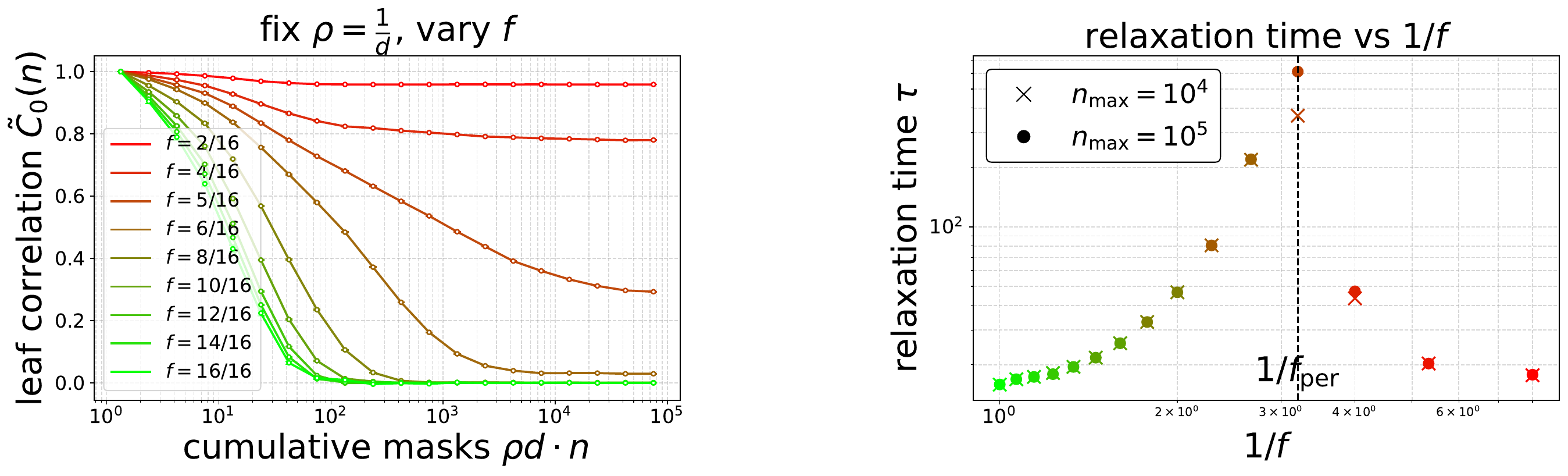}
\caption{
Dynamics of minimal UTMC for the RHM with $L=4$, $s=2$.
\textbf{Left:} Leaf-layer relaxation $\tilde C_0(n)$ versus the cumulative number of masks  $\rho d n=n$, which is simply $n$ here. For large $f$, the correlation decays to the finite-time ergodic
baseline, whereas for smaller $f$ it saturates at a nonzero plateau.
\textbf{Right:} Relaxation time extracted using finite-time plateaus estimated
at two choices of $n_{\max}$. The peak marks the critical slowing-down near the
theoretically predicted percolation threshold $f_{\mathrm{per}}$.
}
\label{fig:rhm_ergodicity}
\end{figure}

\begin{figure}[htbp]
    \centering
    \includegraphics[width=\linewidth]{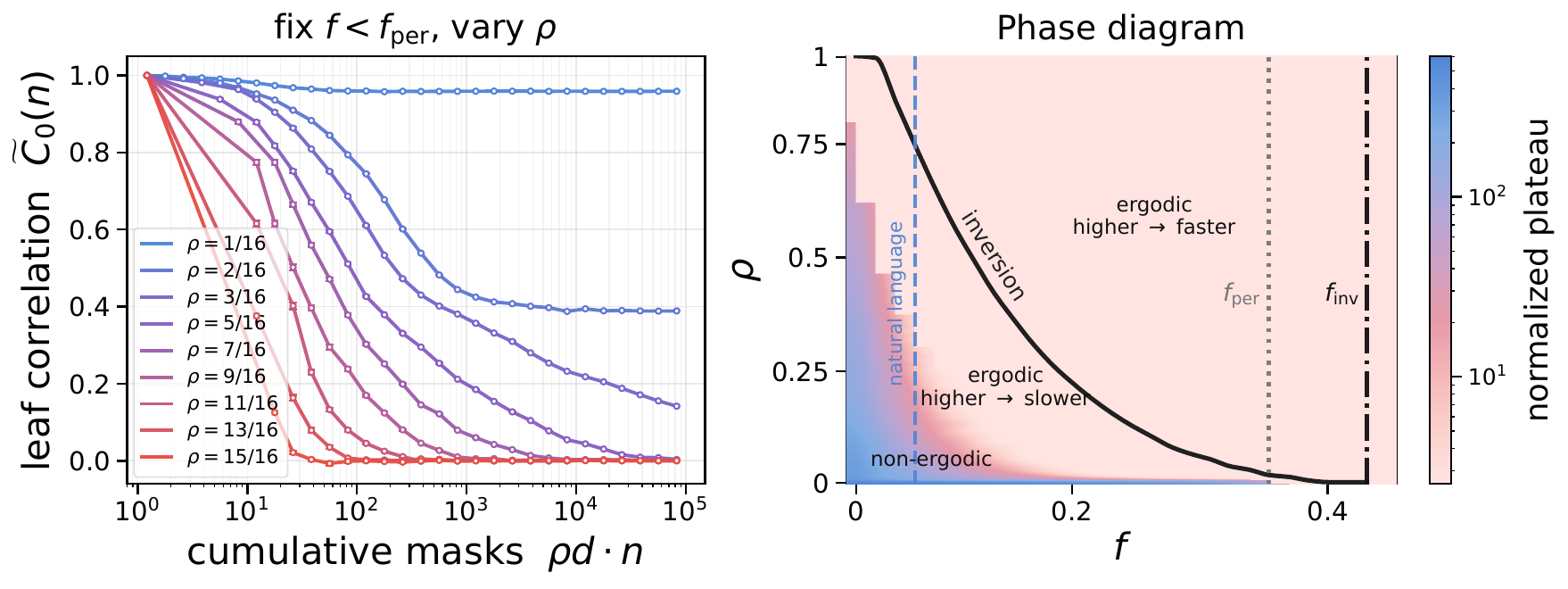}
\caption{
\textbf{Left:} Dynamics of the RHM with $L=4,s=2, f=0.125< f_{\mathrm{per}}$. For these parameters, minimal
UTMC is non-ergodic, but UTMC with larger U-turn steps progressively reduce
the long-time plateau.
\textbf{Right:} Phase diagram for $s=2$, $L=8$, showing the late-time plateau normalized by the standard deviation of the overlap between independent random pairs. Light pink regions are
statistically indistinguishable from the ergodic baseline, while blue regions
retain memory of the initial state. The black curve marks the inversion of
layer-wise relaxation ordering. The values $f_{\mathrm{per}}$ and $f_{\mathrm{inv}}$ are theoretical
predictions for minimal UTMC with $\rho=1/d$.
}
    \label{fig:phase_diagram_ergodicity}
\end{figure}

To track dynamics at different abstraction levels, we measure how the level-$\ell$ representation decorrelates from the initial state after $n$ U-turn steps. Across all modalities, we use
\[
C_\ell(n)=\mathrm{cor}_\ell\!\left(x^{(0)},x^{(n)}\right),
\]
where layer $\ell$ is counted from visible leaves, and $\mathrm{cor}_\ell$ is a modality-specific correlation measure defined in each section.
For the RHM, the natural correlation is the normalized overlap between level-$\ell$ representations:
\[
C_\ell(n)=\frac{1}{s^{L-\ell}}\sum_{i=1}^{s^{L-\ell}}
\delta _{x_\ell^i(0)\,x_\ell^i(n)}.
\]
Since two independent samples can have nonzero overlap, we subtract the independent-sample baseline (formula in App.~\ref{app: rhm setup and observable}), so that $\tilde C_\ell(n)$ indicates loss of memory up to the ergodic baseline. In simulations, plateaus and relaxation times are estimated at finite trajectory length $n_{\max}$. We call $\tilde C_\ell(n_{\max})$ the ``late-time plateau,'' and use the same cutoff when fitting relaxation times. We check robustness by comparing two choices of $n_{\max}$ in Fig.~\ref{fig:rhm_ergodicity}, with additional plateau heat maps (right of Fig.~\ref{fig:phase_diagram_ergodicity}) in App. ~\ref{app:two_plateau_heatmaps}.

\subsection{Minimal U-turns fragment the data distribution}

We first consider minimal U-turn moves, where in each U-turn step one leaf token is masked and
reconstructed, so masking fraction $\rho:=1/d$. The chain then moves on the connectivity graph whose vertices
are valid RHM sentences and whose edges connect sentences differing by one
leaf token.

Figure~\ref{fig:rhm_ergodicity} shows the leaf correlation
$\tilde C_0(n)$ as the rule density $f$ is varied. For dense grammars (large $f$), the
correlation decays to the finite-time ergodic baseline. For sparse grammars, it
instead saturates at a nonzero plateau, indicating that the chain retains memory
of its initial condition. Strikingly, the relaxation time is non-monotonic: it grows as $f$
is decreased from the dense regime, peaks near a critical value, and decreases
again at smaller $f$.

This behavior is the hallmark of a {\it percolation  transition} \cite{stauffer1994introduction} in the local-connectivity
graph. For large $f$, single-token valid flips connect a macroscopic component
of the data support. For small $f$, the graph fragments and a trajectory remains
trapped in the component containing its initial condition. The peak in
relaxation time reflects critical slowing down near the onset of fragmentation.
As the transition is approached from below, fragments become larger and longer to explore, so the chain reaches
its component-specific plateau more slowly. Past the transition, the macroscopic component appears, composed of loosely connected fragments. These become smaller and smaller, and faster to explore,  away from the transition as the macroscopic component becomes more homogeneous.

\subsection{Larger U-turns restore mixing and invert relaxation ordering}

We next increase the masking fraction $\rho$ of each U-turn step. For
$\rho>1/d$, a single U-turn can modify several leaf tokens, thereby
adding longer-range moves to the local connectivity graph. As shown in the left of
Fig.~\ref{fig:phase_diagram_ergodicity}, increasing $\rho$ reduces the
late-time memory plateau and can restore effective ergodicity even in
regimes where minimal U-turn dynamics is fragmented. The value of this late-time plateau  is indicated in the $(f,\rho)$ phase diagram of Fig.~\ref{fig:phase_diagram_ergodicity}, Right panel. 

\begin{figure}[htbp]
    \centering
    \includegraphics[width=\linewidth]{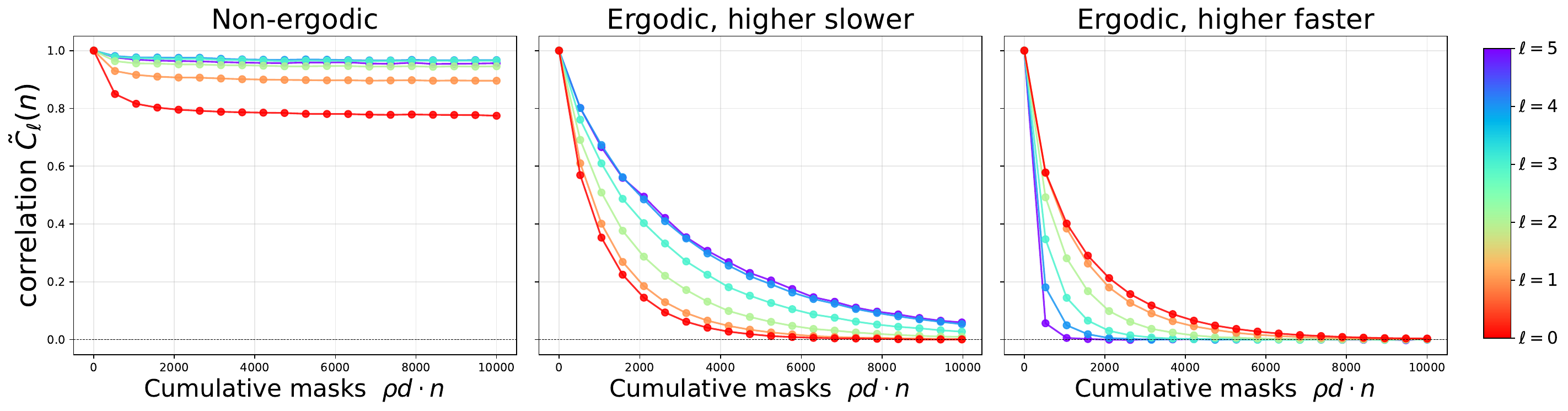}
\caption{
 Layer-wise relaxation across dynamical regimes. $s=2$, $L=8$.
\textbf{Left:} $(f,\rho)=(0.28,\,0.004)$. In the non-ergodic regime, all
layers retain finite memory. \textbf{Middle:} $(f,\rho)=(0.22,\,0.137)$.
In the effectively ergodic, higher-slower regime, all layers eventually
decorrelate, but higher layers relax more slowly. \textbf{Right:}
$(f,\rho)=(0.50,\,0.004)$. In the effectively ergodic, higher-faster
regime, the ordering is inverted and higher layers decorrelate faster.
}
    \label{fig:phase_diagram_layers}
\end{figure}

Layer-wise relaxation reveals a second effect. Depending on $(f,\rho)$,
higher-level latent variables can decorrelate either more slowly or more rapidly
than lower-level latent variables. 
In Fig.~\ref{fig:phase_diagram_layers}, we observe three dynamical
regimes: (i) a non-ergodic regime, in which all layers retain
finite memory of the initial condition; (ii) an effectively
ergodic regime with higher layers relaxing more slowly; and
(iii) an effectively ergodic regime with higher layers relaxing
faster. We quantify this effect by comparing the fitted relaxation times $\tau_{\ell}$ across levels and indicate in the phase diagram of Fig.~\ref{fig:phase_diagram_ergodicity} where the ordering inversion takes place.

\subsection{Theoretical estimates for the percolation and inversion thresholds}
\label{sec:branching_estimates}

%For minimal UTMC, we can theoretically estimate boundaries of such phases. The first scale, $f_{\mathrm{per}}$, concerns
%\emph{connectivity}: whether valid single-token moves form a giant connected
%component in the space of valid sentences. The second scale, $f_{\mathrm{inv}}$,
%concerns \emph{relaxation ordering}: whether higher latent layers decorrelate
%more slowly or more quickly than lower layers.

\paragraph{Known results for the RHM:}
By definition of the RHM, changing a leaf symbol leaves an isolated string valid with probability $f$. There are thus on average $fv$ leaf symbols that are compatible at a given location. The situation is different in the limit $L\to\infty$; see App.~\ref{app:finite threshold} for the finite-$L$ case. The average number of symbols available at a given location was shown to be $1/(1-f)=1+f+f^2+...$ \cite{cagnetta2024towards}.  In this expression, $f^\ell$ indicates the probability that a leaf symbol and $\ell-1$  successive parent symbols change.   

\paragraph{Percolation threshold:}
We argue that the percolation threshold $f_{\mathrm{per}}$ can be computed by considering the mean number of novel flips $n(f)$ that are opening due to a single acceptable flip - see \cite{muller2015marginal} for a review of such arguments in physical systems. If $n(f)>1$, the number of possibilities explodes as more flips are triggered and the connected component is system-spanning. If  $n(f)<1$, the number of available flips dies out and the system is non-ergodic. 

If $i$ and $j$ have a common parent at order $\ell$, flipping $i$ will open a move in $j$ if: (i) $i$ changes its parent latents at least to level $\ell-1$. Conditioned to $i$ being able to flip,  this occurs with probability $(f^{\ell}+ f^{\ell+1}+...)/(f+f^2+...)= f^{\ell-1}$. (ii) Such changes allow for a successive change of $j$ and all its parents to at least level ${\ell-1}$, which occurs with probability $f^\ell + f^{\ell+1}+f^{\ell+2} +..= \frac{ f^{\ell} } {1-f}$. 
Overall, the probability that a new flip is allowed in $j$ is $\frac{ f^{2\ell-1} } {1-f}$.

Summing over all possible positions for $j$, one gets:
\[ n(f) := \sum_{\ell=1}^{L} \left(s^\ell - s^{\ell-1}\right) \frac{ f^{2\ell-1} } {1-f}
\;\xrightarrow[L \to \infty]{}\; \frac{f(s-1)}{(1-sf^2)(1-f) } 
\]
We obtain the percolation threshold by imposing that $n(f_{\mathrm{per}})=1$, leading to:
\begin{equation}
    \frac{f_{per}(s-1)}{(1-sf_{per}^2)(1-f_{per}) } 
    =1 , \qquad   f_{\mathrm{per}} 
    \simeq
    \frac{1}{s}
    -
    \frac{1}{s^2}
    +
    O(s^{-3}) .
\end{equation}

\paragraph{Layer-ordering inversion threshold $f_{\mathrm{inv}}$:}
%The second threshold concerns not connectivity, but the ordering of relaxation
%times across layers.
To estimate the inversion threshold $f_{\mathrm{inv}}$, we consider which fraction of latent variables is affected on average by flipping a fraction of their children nodes. There are two competing effects. (i) There
are fewer parents than children: level $\ell+1$ contains a factor $s$ fewer
variables than level $\ell$; so that $s$ flips of the latter can affect the parent. This effect favors faster decorrelation of higher layers. (ii) A flip at any level triggers a flip of its parent with probability $f$.
We predict that the ordering changes when these two effects balance:
\begin{equation}
    s f_{\mathrm{inv}} = 1,
    \qquad
    f_{\mathrm{inv}} = \frac{1}{s}.
\end{equation}

Finite-\(L\) corrections to \(f_{\mathrm{per}}\) and \(f_{\mathrm{inv}}\) are derived in App.~\ref{app:finite threshold}; these finite-size estimates are used in Fig.~\ref{fig:rhm_ergodicity} and Fig.~\ref{fig:phase_diagram_ergodicity}.

\section{U-turn chain on natural language}
\label{sec:language}

We next test whether the theoretical predictions from Section~\ref{sec:theory} carry over to natural language. %In the RHM, U-turn dynamics revealed three qualitative signatures: slow relaxation of high-level variables in the small-step regime, an inversion of layer-wise relaxation ordering at sufficiently large U-turn magnitude. We now ask whether analogous signatures appear when U-turn chains are generated by a masked diffusion language model and the evolving text is probed using the internal representations of an autoregressive LLM.

\paragraph{Experimental setup.}

We use Dolma \cite{soldaini-etal-2024-dolma} as the natural-language dataset. We sample 256 text sequences and crop each sequence to token length $d=128$. UTMC are generated using the LLaDA 7B MoE base model \cite{zhu2025llada} as the masked diffusion language model. As a proxy for how low and high-level features of text evolve along the chain, we consider the residual-stream activations of Mistral 7B \cite{jiang2023mistral7b}. In main text we show result for all layers, while in App.~\ref{app:language_single_step} we show plot restricted to first 15 layers. This choice is motivated by prior work showing that LLMs often self-organize into an encoder-decoder design, where intermediate layers contain the most abstract representations~\cite{caucheteux2022brains,caucheteux2022deep,cheng2025emergencehighdimensionalabstractionphase}. Unless otherwise stated, reported curves are averaged over 256 initial texts and 64 independent UTMC per text. Additional details of experiments are in App.~\ref{app:language_implementation_details}. In  App.~\ref{app:dream}, we report additional experiments using different masked diffusion language models, which qualitatively agree with the main-text results.

Note that for the diffusion model considered, we found that long U-turn trajectories accumulate errors from the masked diffusion model. The generated text eventually becomes degraded or nonsensical, as indicated by the perplexity measurement shown in App.~\ref{app:perplexity}. To avoid significant corruption we limit the U-turn trajectory lengths considered, up to about $300$ cumulative masked tokens, where afterwards perplexity explodes.   

\begin{figure}[htbp]
    \centering
    \includegraphics[width=\linewidth]{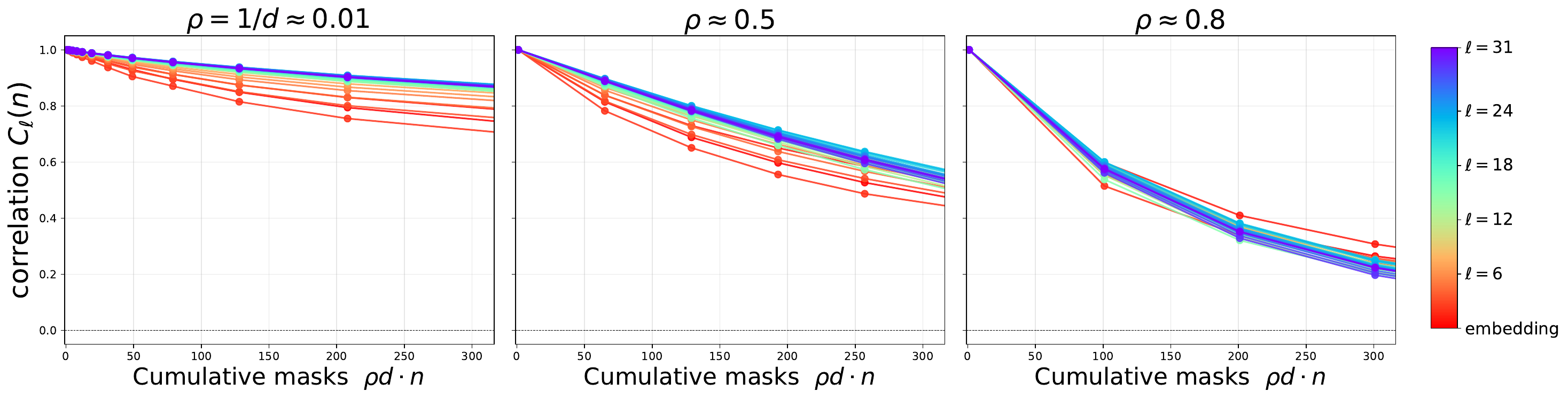}
\caption{
\textbf{Layer-wise latent correlation and ordering inversion in text.} \textbf{Left:} For minimal U-turn steps, all layers relax slowly, with deeper layers in the early-to-intermediate range retaining memory of the initial text for longer.
\textbf{Middle:} Increasing the masking fraction accelerates decorrelation across layers.
\textbf{Right:} At large masking fraction, the layer ordering inverts: deeper representations decorrelate faster than earlier layers, consistent with the high-$\rho$ regime predicted by the RHM. More choices of $\rho$ are reported in App.~\ref{app:language_sequential_multiple_panels}.
}
    \label{fig:text_layer_inversion}
\end{figure}

\paragraph{Observables.}
For each U-turn step $n$, we feed both initial text and the U-turned text into the probe LLM. At layer $\ell$, we extract the residual-stream activation and mean-pool over token positions, obtaining vectors $\vec{x}_\ell(0)$ and $\vec{x}_\ell(n)$. We then subtract a layer-dependent baseline activation $\vec{\mu}_\ell$, computed by averaging activations over highly corrupted U-turn samples with masking fraction $\rho \approx 1$. This subtraction removes layer-specific common components and centers the representation relative to the highly mixed regime of the U-turn dynamics. The layer-wise correlation is defined as
\[
C_\ell(n)
=
\cos\!\left(
\vec{x}_\ell(0)-\vec{\mu}_\ell,\,
\vec{x}_\ell(n)-\vec{\mu}_\ell
\right).
\]
This observable measures how much the representation at layer $\ell$ retains memory of the initial text after $n$ U-turn steps.

\paragraph{Results.} Our main results are remarkably consistent with theoretical predictions: (i) For minimal U-turn dynamics, mixing is poor and relaxation is very slow as shown in the Left of Fig.~\ref{fig:text_layer_inversion}. (ii) Mixing improves considerably as the fraction of masking at each U-turn  increases, as apparent in the Middle and Right of Fig.~\ref{fig:text_layer_inversion}. (iii) Deeper  layers relax more slowly than earlier layers, as shown in the Left and Middle of Fig.~\ref{fig:text_layer_inversion}, except at larger masking where this trend  inverts as shown on the Right of this figure. 

We use the masking level where this inversion takes place to locate Dolma approximately in the phase diagram of Fig.~\ref{fig:phase_diagram_ergodicity}.

\section{U-turn chain on images}
\label{sec:images}

In this section, we investigate whether the layer-wise dynamical signatures from Section~\ref{sec:theory} also emerge in natural images. %We ask whether an analogous pattern appears when image U-turn chains are generated by a pretrained diffusion model and probed using the internal representations of a convolutional classifier.

\paragraph{Experimental setup.}
All image experiments use ImageNet (ILSVRC2012)~\cite{deng2009imagenet} and a pretrained 256$\times$256 diffusion model implemented in the \texttt{guided-diffusion} codebase~\cite{dhariwal2021diffusion}. Unless stated otherwise, trajectories are initialized from ImageNet validation images rather than from noise. Images are represented as RGB tensors scaled to $[-1,1]$. To probe visual representations along the chain, we use a ConvNeXt-Base classifier pretrained on ImageNet~\cite{liu2022convnext}. 

\begin{figure}[htbp]
    \centering
    \includegraphics[width=\linewidth]{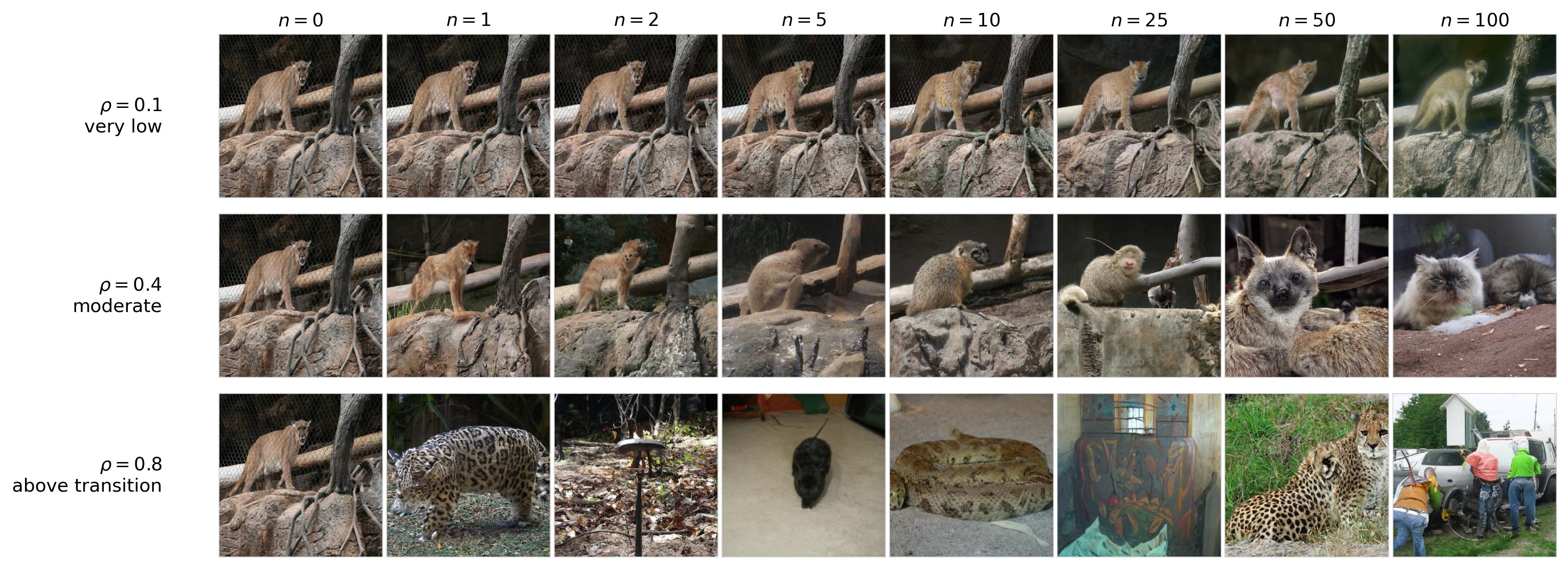}
    \caption{\textbf{Representative sequential U-turn trajectories.} Rows use the same initial ImageNet validation example and trajectory index from the latent-analysis dataset, with fixed noise fractions $\rho=0.1,0.4,0.8$. Columns show the sequential U-turn index $n$. The qualitative drift accelerates as the per-step noise increases, matching the quantitative collapse of latent correlation.}
    \label{fig:app_image_sequential_montage}
\end{figure}

We iterate image U-turns at diffusion times
$t\in\{100,400,800\}$ under the standard $T=1000$ noise schedule, corresponding to
$\rho=t/T\in\{0.1,0.4,0.8\}$. Each trajectory is initialized from an ImageNet validation image and run for $N=100$ U-turn steps. Representative examples of trajectories are shown in Fig.~\ref{fig:app_image_sequential_montage}.

\begin{figure}[htbp!]
    \centering
    \includegraphics[width=\linewidth]{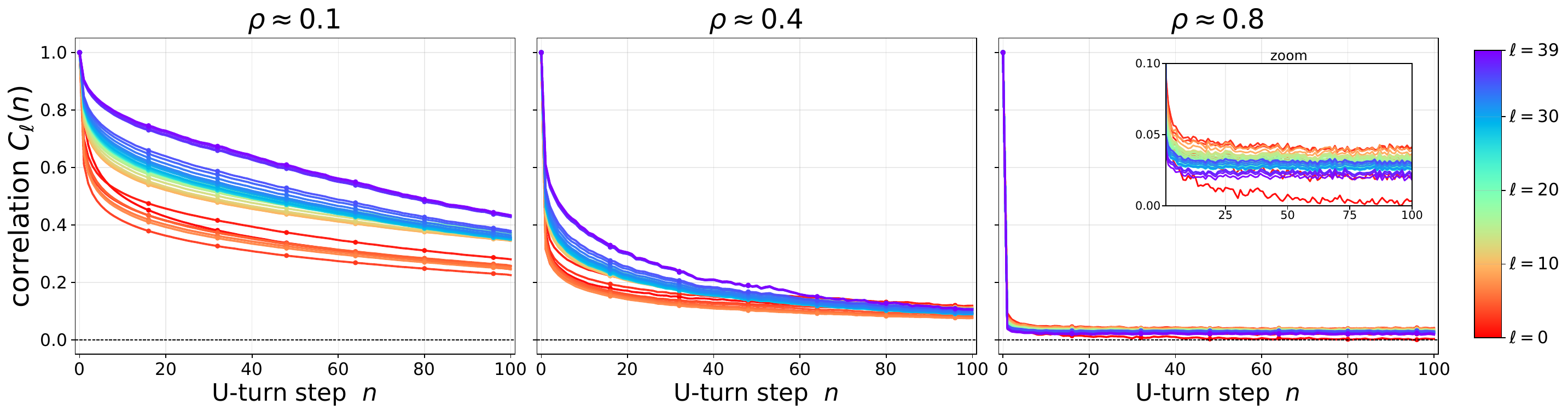}
    \caption{\textbf{Layer-wise latent correlation and ordering inversion in images.} Cosine correlation $C_\ell(n)$ of ConvNeXt feature activations between the initial image and sequential U-turn states, averaged over the 20-image ImageNet validation images. Colors denote ConvNeXt feature depth, from early layers (red) to deep layers (purple); the classifier head is excluded. \textbf{Left:} At small U-turn magnitude, $\rho\simeq 0.1$, deeper visual representations retain memory of the initial image for longer than early layers. \textbf{Middle:} Near the transition, $\rho\simeq 0.4$, correlations decay more rapidly and the layer ordering narrows. \textbf{Right:} At high noise, $\rho\simeq 0.8$, the ordering inverts, with deep layers decorrelating faster than early layers; the inset zooms into $0\leq C_\ell(n)\leq 0.1$ to resolve this separation.}
    \label{fig:image_latent_persistence}
\end{figure}

\paragraph{Observables.}
At each step $n$, we probe the image $x^{(n)}$ with ConvNeXt-Base and record the activation
$\phi_\ell(x^{(n)})$ at layer $\ell$. Then we whiten each layer using fixed ImageNet statistics, $ \tilde\phi_\ell =\frac{\phi_\ell-\mu_\ell}{\sqrt{\sigma_\ell^2}}$,
with $(\mu_\ell,\sigma_\ell^2)$ estimated from 100 ImageNet images. Correlation is defined
\[
C_\ell(n)=
\cos\!\big(
\tilde\phi_\ell(x^{(n)}),
\tilde\phi_\ell(x^{(0)})
\big).
\]

\paragraph{Results.}
Figure~\ref{fig:image_latent_persistence} shows results  averaged over a 20-image dataset, first within each image and then across images. Again, agreement with theory is remarkable: (i) at small U-turn magnitude, early feature layers drift rapidly while deeper layers retain high correlation for much longer. (ii) Increasing $\rho$ accelerates decorrelation across all layers. (iii) At sufficiently large $\rho$, the ordering inverts: deep features decorrelate slightly faster than the early layers. Additional qualitative reconstructions, integrated persistence summaries, single-step diagnostics, and  details are reported in App.~\ref{app:image_diffusion}.

\section{Conclusion}

We introduced U-turn Markov chains (UTMC), obtained by iterating short
forward--backward diffusion steps; for an exact diffusion model, the
U-turn kernel satisfies detailed balance with respect to the data
distribution. In the Random Hierarchy Model we identified two
dynamical signatures of data geometry: (i) a connectivity transition
under which minimal-step dynamics fragments the state space, with
ergodicity restored by larger U-turns; and (ii) a layer-dependent
relaxation in which higher-level variables relax more slowly than
lower-level ones in the small-step regime, an ordering that inverts at
sufficiently large U-turn magnitude. Both signatures appear in real
data: deep LLM representations decorrelate much more slowly than early
ones under minimal U-turns on natural language, and the layer-ordering
inversion together with the dynamical-susceptibility peak occur only
at large masking fraction; ConvNeXt features show an analogous
depth-ordered relaxation on natural images. Mapped onto the RHM phase
diagram, these signatures locate natural language  in the
non-ergodic region, where minimal U-turn dynamics is strongly
constrained and confined to a restricted component of the learned data
manifold.

\textbf{Future directions.}
The phase diagram has a direct consequence for using UTMC as a
Monte-Carlo tool. Ergodicity can be restored by choosing U-turn moves
large enough to bridge the disconnected components of the
local-connectivity graph, but the price is that such moves are coarse
in data space: they are ill-suited to fine-tuning the optima of an
auxiliary energy $H(x)$, or to exploring the vicinity of a particular
sample of interest. Small steps explore locally but cannot leave their
initial component, while large steps mix globally but lack the
resolution required for delicate optimization. Quantifying this
trade-off empirically---for instance, by steering images toward
attributes by taking $H$ as the output of a classifier, or biasing text
toward a target sentiment by taking $H$ to be a sentiment score---is a
natural next direction, and would turn the diagnostic dynamics studied
here into a practical tool for controlled generation.

A second direction concerns interpretability. UTMC offer a new lens on
the internal representations of large language models: tracking how
specific syntactic and semantic features---grammatical agreement,
named entities, topic, sentiment---evolve along a U-turn chain, and
relating their relaxation times to those of LLM hidden representations
at different depths, would tie a generative dynamical probe to the
standard layer-by-layer probing
methodology~\cite{conneau2018cram,tenney2019bert,manning2020emergent,kulmizev2022schrodinger}.
This perspective could clarify which abstraction levels are stable
under local diffusion-model perturbations and which are most plastic,
and whether the depth at which a feature is encoded in an LLM
correlates with its relaxation time under UTMC.

\paragraph{Limitations.}
Our theoretical analysis is carried out in the Random Hierarchy Model with an exact Bayes-optimal denoiser, whereas real diffusion models introduce approximation error that can accumulate along long U-turn trajectories. It would be interesting to study theoretically how errors accumulate in synthetic languages where the ground truth is known. %This is especially visible in the language experiments, where repeated masked denoising can eventually degrade text quality, limiting the trajectory lengths over which mixing can be reliably measured. Exploring whether larger or more accurate diffusion language models, such as LLaDA 2.0\cite{bie2025llada20scalingdiffusionlanguage}, mitigate this degradation is an interesting direction for future work.

Moreover, the RHM is a deliberately minimal proxy for hierarchical compositional
data: its tree topology is regular (fixed depth $L$, branching $s$),
the grammar is non-recursive and strictly context-free, and within
each parent the allowed production rules are activated with equal
probability rather than following the heavy-tailed (e.g.\ Zipf)
distribution observed in natural language. Closing the gap with real
data calls for (i) random or sample-dependent tree topologies as in \cite{parley2026deepnetworkslearnparse,allenzhu2024cfg, zhao2023transformers, garnierbrun2024transformers},
(ii) recursive grammars \cite{degiuli2019random,schulz2025unraveling}, (iii) context-dependent rules, and
(iv) Zipf-distributed rule probabilities \cite{cagnetta2025learningcurvestheoryhierarchically}. We expect such extensions to
shift the percolation and layer-inversion thresholds $f_{\rm per}$ and
$f_{\rm inv}$ while leaving the qualitative non-ergodic, hierarchically
stratified picture intact.

\paragraph{Acknowledgements}
We thank all members of PCSL, Marco Baroni, and Eric Vanden-Eijden for useful discussion. This work was supported by the Simons Foundation through the Simons Collaboration
on the Physics of Learning and Neural Computation (Award
ID: SFI-MPS-POL00012574-05), PIs  Wyart. D. J.
Korchinski acknowledges financial support from the Natural Sciences and Engineering Research Council of Canada (NSERC PDF - 587940 - 2024).

\bibliographystyle{unsrtnat}

\bibliography{bibib}
% \bibliography{NeurIPS2026/bib}

%%%%%%%%%%%%%%%%%%%%%%%%%%%%%%%%%%%%%%%%%%%%%%%%%%%%%%%%%%%%
%                         APPENDIX
%%%%%%%%%%%%%%%%%%%%%%%%%%%%%%%%%%%%%%%%%%%%%%%%%%%%%%%%%%%%

\clearpage
\appendix

%\section{Dynamical susceptibility of RHM and natural language}
%\begin{figure}[htbp]
%    \centering
%    \includegraphics[width=\linewidth]{figs/rhm_montecarlo_fig_susceptibility.pdf}
%\caption{
%Dynamical susceptibility identifies the characteristic U-turn scale.
%Single-step dynamical susceptibility $\chi_4(\rho)$ as a function of masking fraction.
%\textbf{Left:} In the RHM with $f<f_{\mathrm{per}}$, susceptibility peaks at a large $\rho$, indicating the scale at which U-turns begin to induce collective changes in higher-level variables.
%\textbf{Right:} Natural language exhibits a similar peak at large masking fraction, suggesting that substantial corruption is required to perturb high-level semantic structure.
%}
%    \label{fig:placeholder}
%\end{figure}

% \section{Stationary distribution of the U-turn chain}
% \label{Appendix: stationary distribution}
% \paragraph{Proposition 1.}
% Assume the diffusion model is exact, i.e.\ $p_\theta(x \mid x_t)=p(x \mid x_t)$. Then the U-turn kernel satisfies detailed balance with respect to $P(x)$:
% $
% U_t(x',x)P(x)=U_t(x,x')P(x').$
% If the resulting Markov chain is ergodic, its stationary distribution is $P(x)$.

% \paragraph{Proof (sketch).} 
% Using Bayes' rule $p(x'|x_t)=\frac{q(x_t|x')P(x')}{p(x_t)}$
% in the definition of $U_t$ yields
% \[
% U_t(x',x)P(x)
% =
% \int
% \frac{q(x_t|x)q(x_t|x')}
%      {p(x_t)}
% P(x)P(x')\,dx_t,
% \]
% which is symmetric in $x$ and $x'$. $\square$

\section{Plateau heatmaps for different choices of $n_{\max}$}
\label{app:two_plateau_heatmaps}

\begin{figure}[htbp]
    \centering
    \includegraphics[width=\linewidth]{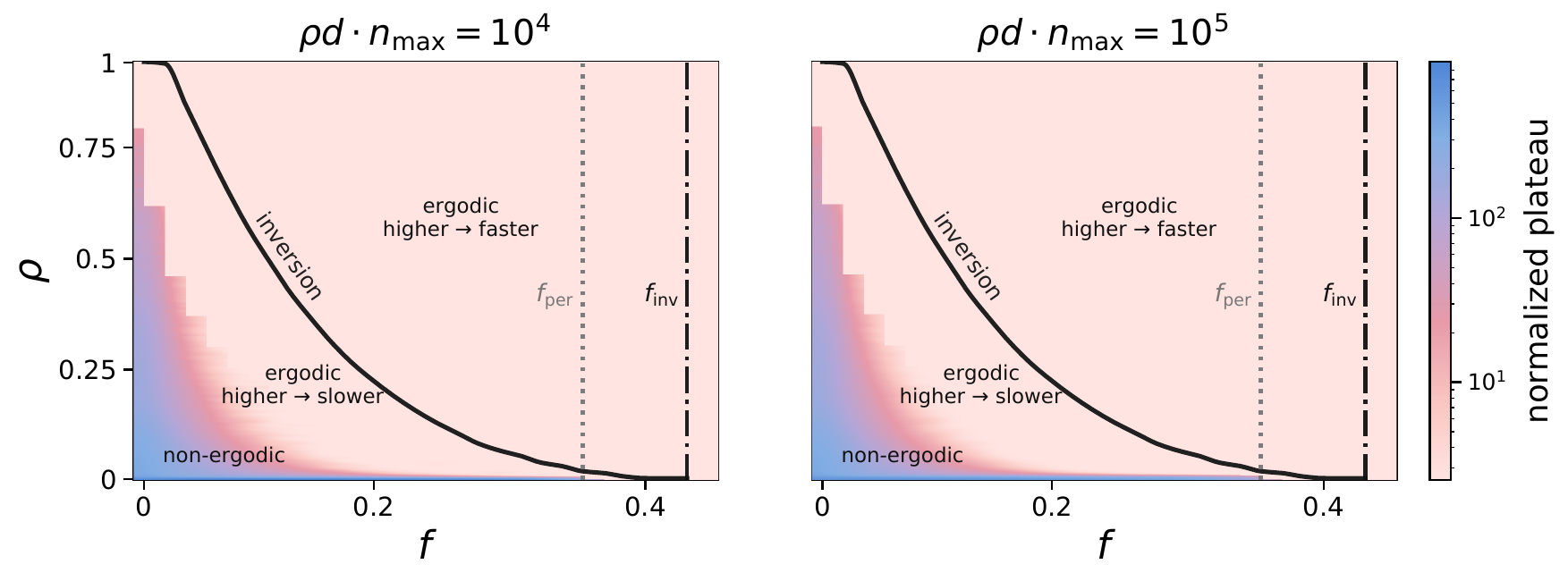}
    \caption{
    Plateau heatmaps for the RHM with $s=2$ and $L=8$, computed using two trajectory lengths.
    \textbf{Left:} $\rho d \cdot n_{\max}=10^4$.
    \textbf{Right:} $\rho d \cdot n_{\max}=10^5$.
    The qualitative structure of the phase diagram is stable across these two choices, indicating that the observed non-ergodic and effectively ergodic regions are not artifacts of a single finite U-turn step cutoff.
    }
    \label{fig:rhm_two_plateau_heatmaps}
\end{figure}

\section{Additional language diffusion results}
\label{app:language_diffusion}

\subsection{Correlation decay of UTMC at varying masking fraction $\rho$}
\label{app:language_sequential_multiple_panels}

\begin{figure}[htbp]
    \centering
    \includegraphics[width=\linewidth]{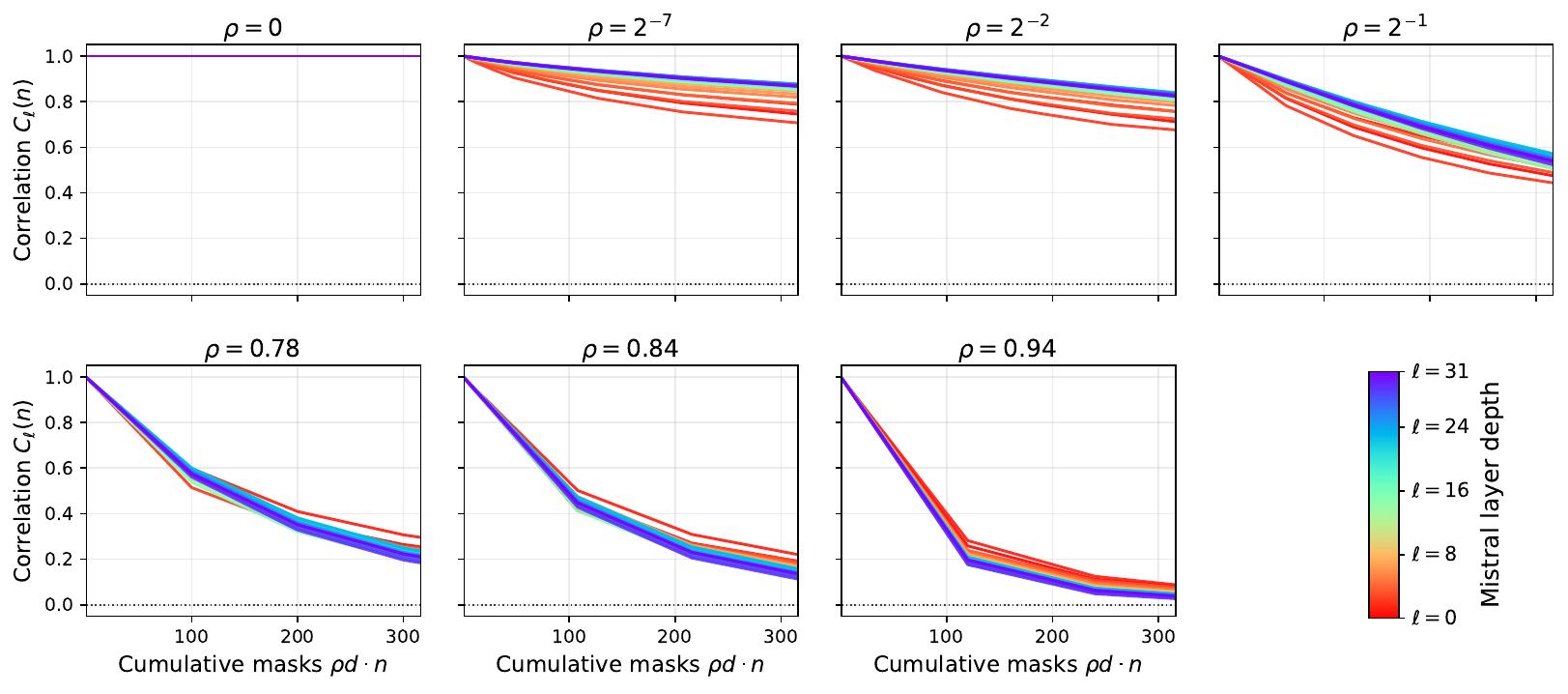}
    \caption{
    Decay of Mistral representation correlations along sequential language U-turn chains for several masking fractions $\rho$.
    Increasing $\rho$ accelerates decorrelation across layers and eventually inverts layer ordering, consistent with the main-text observation.
    }
    \label{fig:sequential_uturn_mistral_selected_rhos_panels_paper_style}
\end{figure}

\subsection{Single-step analysis}
\label{app:language_single_step}

Long U-turn trajectories can eventually degrade text quality, as shown in App.~\ref{app:perplexity}. This makes it difficult to compare fitted relaxation times across masking fractions $\rho$ over very long trajectories. As a complementary diagnostic, we therefore examine the effect of a single U-turn step as $\rho$ is varied. This single-step analysis directly measures which layers are most sensitive to a U-turn perturbation at each masking level, and provides a cleaner way to locate the layer-ordering inversion.

\begin{figure}[htbp]
    \centering
    \begin{subfigure}{0.48\linewidth}
        \centering
        \includegraphics[width=\linewidth]{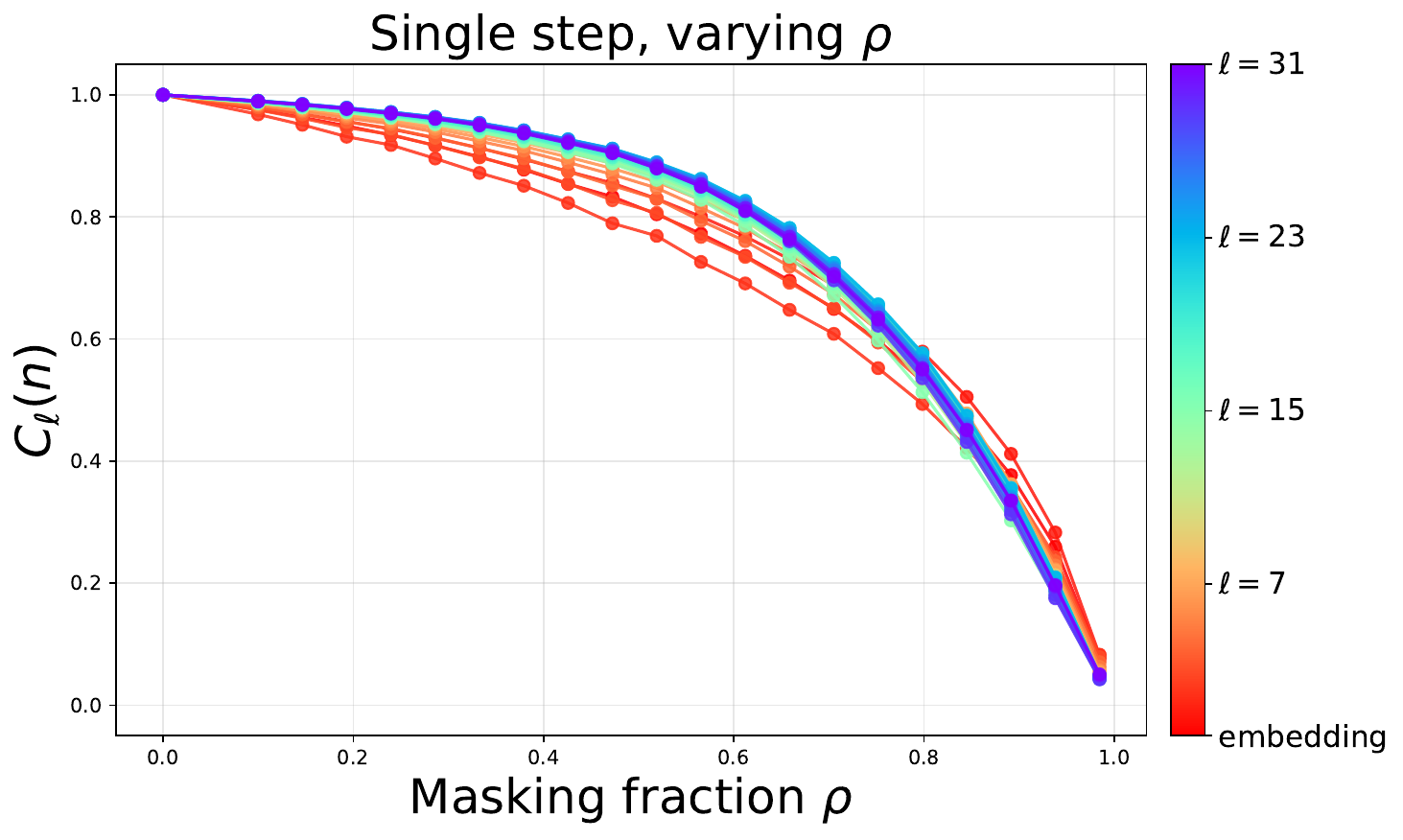}
    \end{subfigure}
    \begin{subfigure}{0.48\linewidth}
        \centering
        \includegraphics[width=\linewidth]{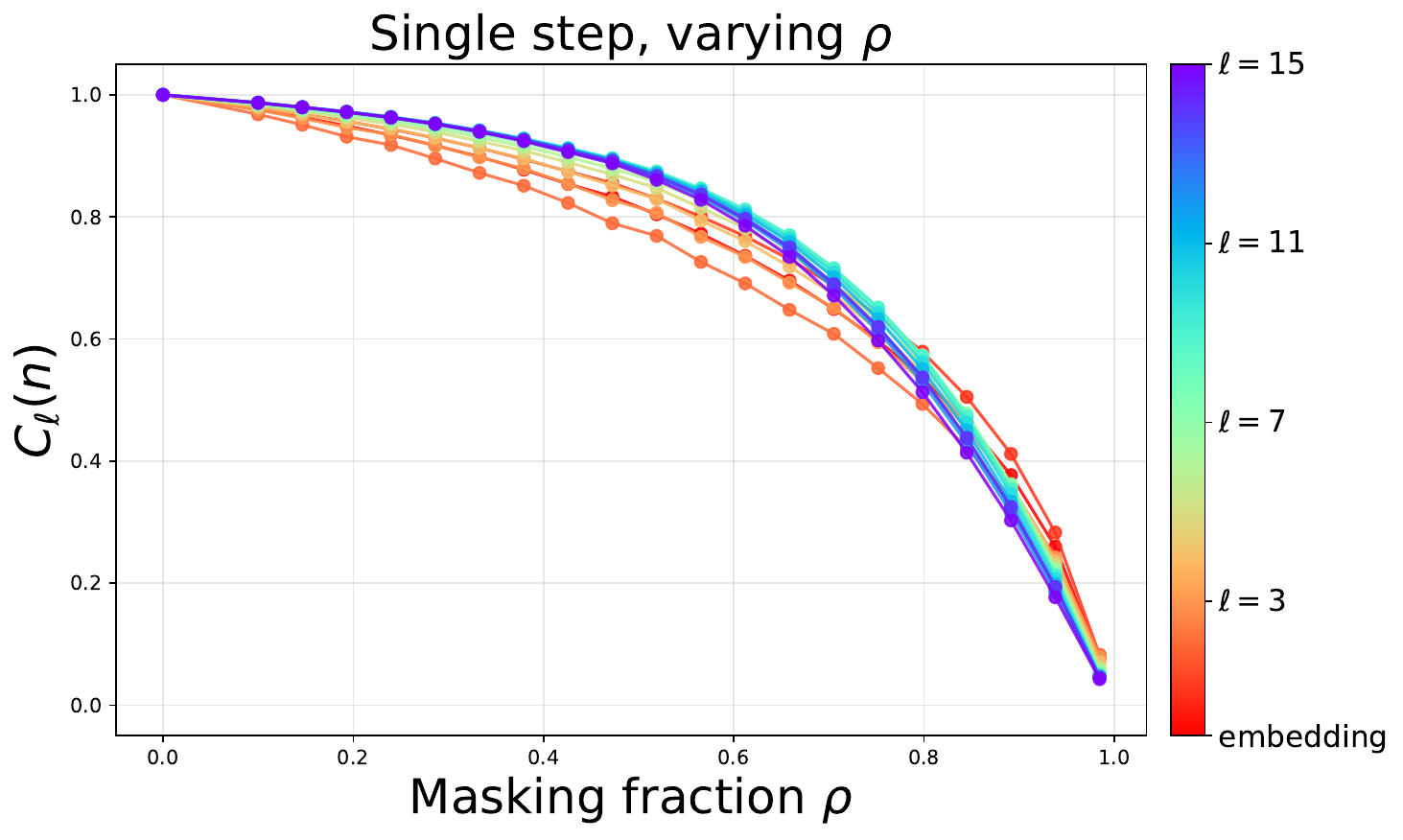}
    \end{subfigure}
    \caption{
    Centered cosine similarity of Mistral representations after a single language U-turn step, as a function of masking fraction $\rho$.
    \textbf{Left:} all Mistral layers.
    \textbf{Right:} the first 15 layers only.
    In both views, the layer ordering inverts around $\rho \approx 0.8$: at small $\rho$, deeper/intermediate representations are more stable, while at large $\rho$ they decorrelate more rapidly.
    }
    \label{fig:single_uturn_mistral_first_15_layers}
\end{figure}

\subsection{Perplexity of U-turned text}
\label{app:perplexity}

We monitor the quality of generated text using Mistral 7B perplexity. This provides a practical diagnostic for when repeated U-turn denoising begins to produce degraded or nonsensical text.

\begin{figure}[htbp]
    \centering
    \begin{subfigure}{0.48\linewidth}
        \centering
        \includegraphics[width=\linewidth]{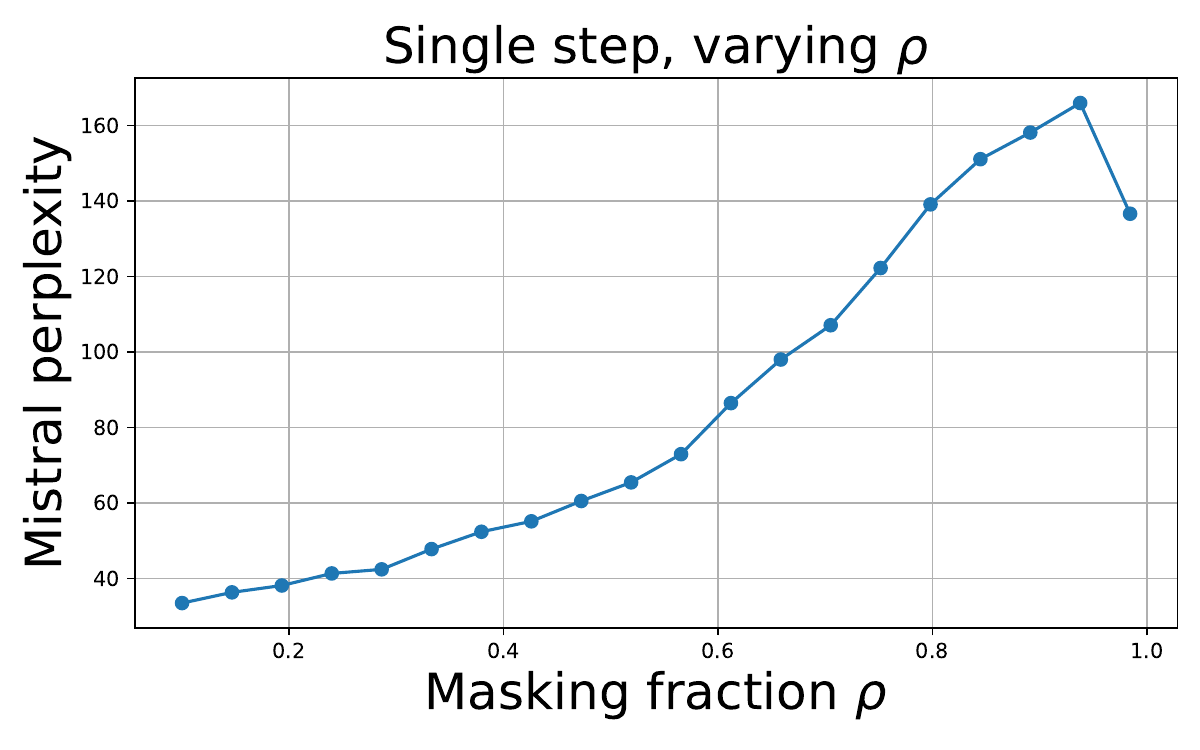}
        \label{fig:single_uturn_ppl}
    \end{subfigure}
    \begin{subfigure}{0.48\linewidth}
        \centering
        \includegraphics[width=\linewidth]{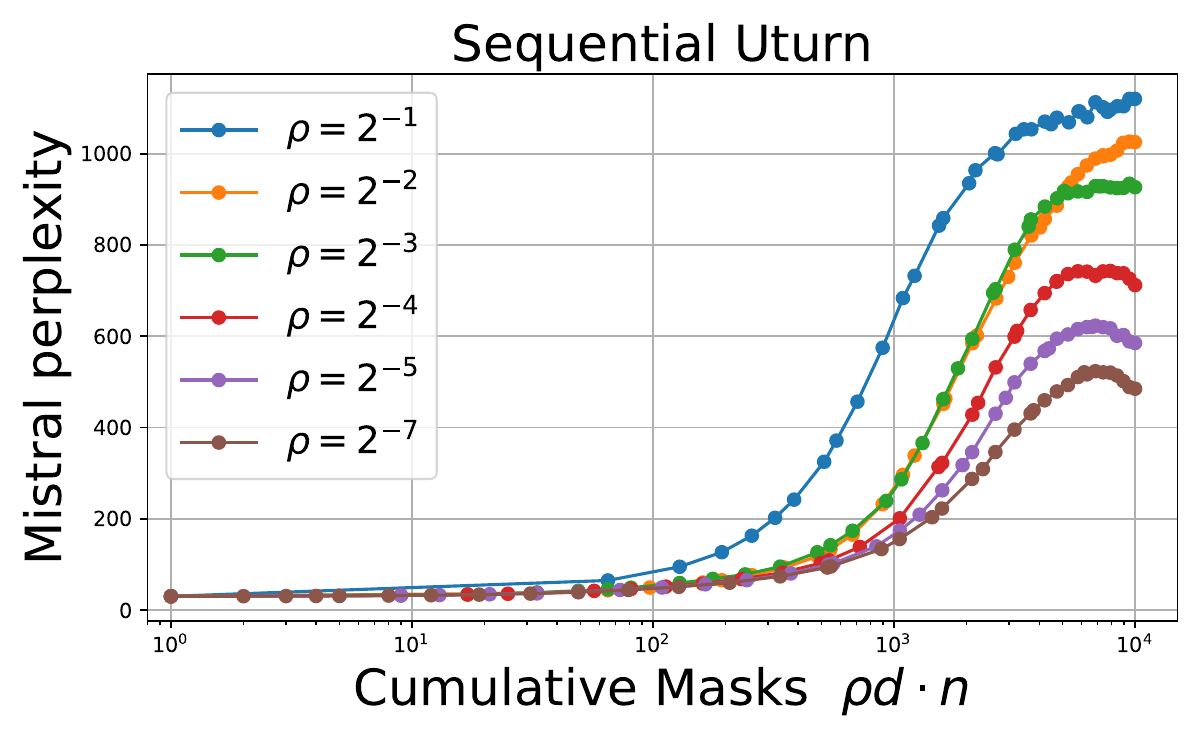}
        \label{fig:sequential_uturn_ppl}
    \end{subfigure}
    \caption{
    Perplexity of text produced by language U-turns, measured using Mistral 7B.
    \textbf{Left:} perplexity after a single U-turn step as the masking fraction $\rho$ is varied.
    \textbf{Right:} perplexity along sequential U-turn trajectories, plotted against the cumulative number of masked tokens $\rho d \cdot n$.
    Perplexity begins to increase sharply around $\rho d \cdot n \approx 300$, indicating that very long sequential U-turn trajectories accumulate denoising errors.
    }
    \label{fig:uturn_ppl_by_rho}
\end{figure}

\subsection{Implementation details}
\label{app:language_implementation_details}

All language U-turn experiments use the \texttt{LLaDA-MoE-7B-A1B-Base} masked diffusion language model on Dolma text. We sample 256 example paragraphs from \texttt{allenai/dolma} with configuration \texttt{v1\_6-sample}, normalize whitespace, and keep clean 128-token prefixes under the LLaDA tokenizer. When constructing the starting set, we exclude examples containing newlines, code-like symbols, CJK characters, or other non-prose artifacts.

For a U-turn with masking fraction $\rho$, the forward step masks $\mathrm{round}(\rho d)$ token positions in the current sequence, where $d=128$. The backward step applies the standard LLaDA iterative masked-token denoising procedure, initialized from the partially masked sequence. Unless otherwise stated, sampling uses temperature $1.0$ and random unmasking.

To probe representations, saved LLaDA token sequences are decoded to text and passed through \texttt{mistralai/Mistral-7B-v0.1}. At each Mistral layer $\ell$, residual-stream activations are mean-pooled over token positions and centered by a layer-wise baseline $\mu_\ell$, computed from highly corrupted U-turn samples with $\rho \approx 1$. The language correlation is defined as
\[
C_\ell(n)
=
\cos\!\left(
\bar h_\ell(x^{(0)})-\mu_\ell,\,
\bar h_\ell(x^{(n)})-\mu_\ell
\right).
\]
Curves are averaged over initial texts and independent U-turn trajectories.

\subsection{Robustness of the natural language results}
\label{app:dream}

To test whether the natural-language layer-ordering inversion is specific to LLaDA, we repeat the single-step analysis using a different masked diffusion language model, Dream 7B v0 Base~\cite{ye2025dream}. Starting from the same dataset, we observe a qualitatively similar inversion of representation decay across Mistral layers, consistent with the behavior shown in App.~\ref{app:language_single_step}.

\begin{figure}[htbp]
    \centering
    \begin{subfigure}{0.48\linewidth}
        \centering
        \includegraphics[width=\linewidth]{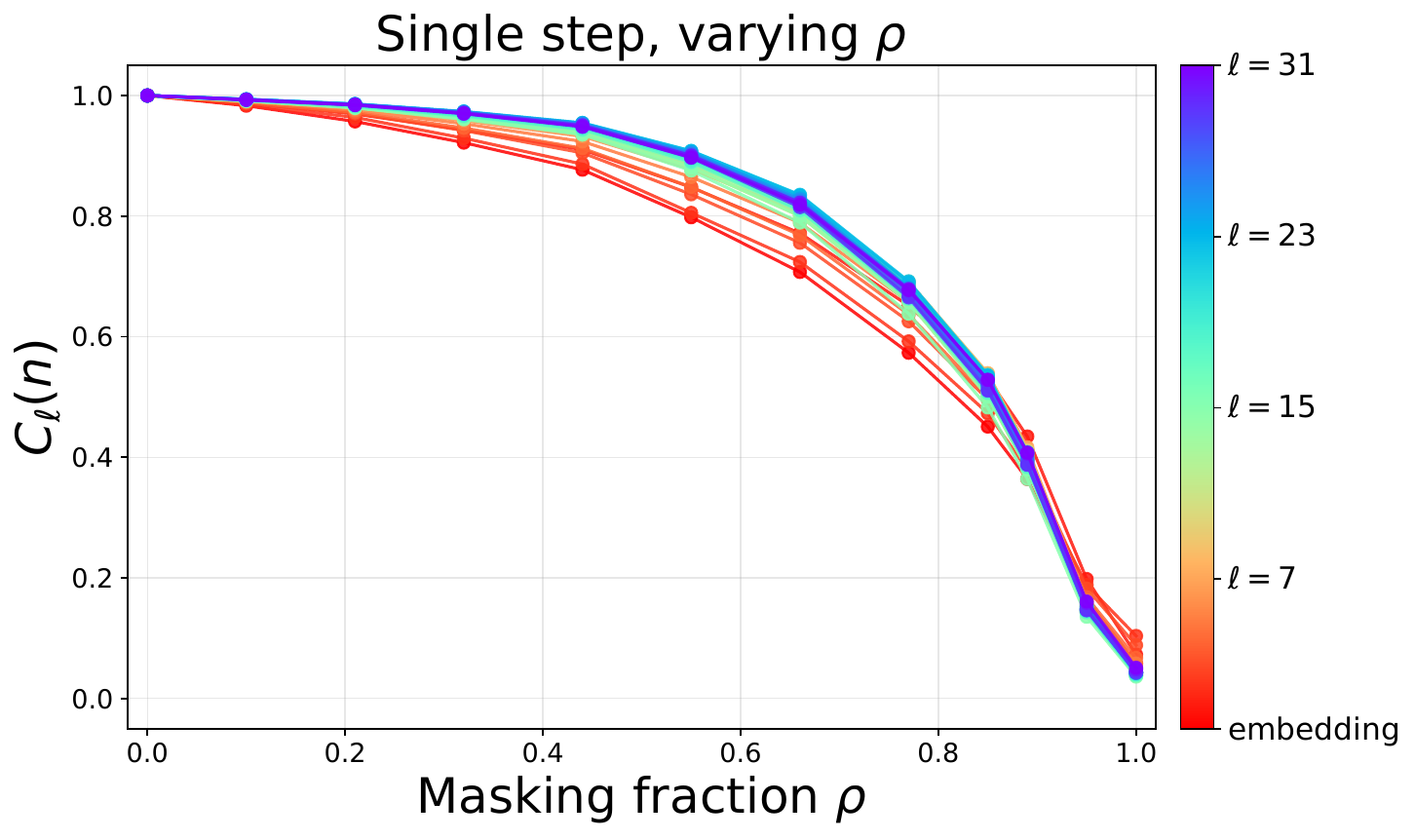}
    \end{subfigure}
    \begin{subfigure}{0.48\linewidth}
        \centering
        \includegraphics[width=\linewidth]{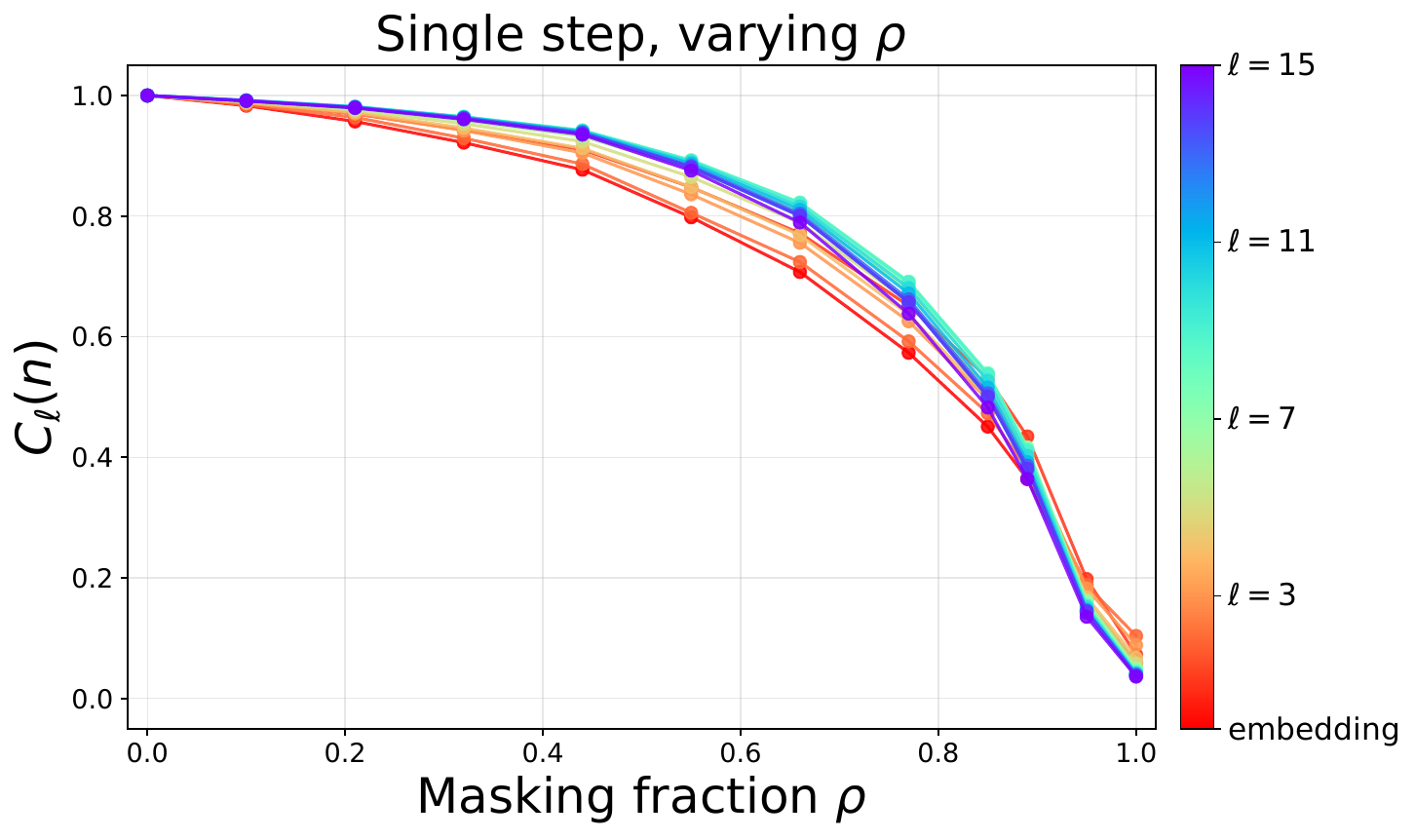}
    \end{subfigure}
    \caption{
    Centered cosine similarity of Mistral representations after a single language U-turn step using Dream 7B v0 Base as the masked diffusion model, as a function of masking fraction $\rho$.
    \textbf{Left:} all Mistral layers.
    \textbf{Right:} the first 15 layers only.
    As in the LLaDA experiments, intermediate and deeper layers become less stable relative to earlier layers at large masking fraction, indicating a similar layer-ordering inversion.
    }
    \label{fig:single_uturn_mistral_first_15_layers}
\end{figure}

% \begin{figure}[htbp]
%     \centering
%     \includegraphics[width=0.5\linewidth]{figs/plot_first_uturn_vs_mf_no_rescale.pdf}
%     \caption{
%     Centered cosine similarity of Mistral representations after a single U-turn step using Dream 7B v0 Base as the masked diffusion model.
%     As in the LLaDA experiments, intermediate and deeper layers become less stable relative to earlier layers at large masking fraction, indicating a similar layer-ordering inversion.
%     }
%     \label{fig:single_uturn_dream}
% \end{figure}

Dolma is included in the training mixture of Dream 7B v0 Base, which makes this experiment a useful robustness check: the inversion persists even when the diffusion model is evaluated on data drawn from a distribution it was trained on. In contrast, the training data for LLaDA 7B are not publicly specified.

\section{Additional Image Diffusion Results}
\label{app:image_diffusion}

This appendix collects additional image experiments supporting \Cref{sec:images}. Unless stated otherwise, the U-turn map is the same as in the main text: starting from an image $x$, we sample $x_t\sim q(x_t\mid x)$ at diffusion time $t$, then run the ancestral reverse sampler for $t$ steps to obtain a reconstructed image. We report the noise fraction as $\rho=t/T$ with $T=1000$.

% \subsection{Qualitative sequential U-turn sweeps}

% \Cref{fig:app_image_sequential_montage} shows representative sequential U-turn trajectories from the same high-noise ImageNet pilot used for the latent statistics below. Each row fixes the per-step noise level and shows the image after increasing numbers of U-turns, from the initial image at $n=0$ to $n=100$. At very small noise the trajectory remains near the starting image for many steps, at moderate noise it drifts through related local configurations, and above the transition it rapidly moves to different regions of the image distribution.

% \begin{figure}[htbp]
%     \centering
%     \includegraphics[width=\linewidth]{figs/image_appendix_sequential_uturn_montage.png}
%     \caption{\textbf{Representative sequential U-turn trajectories.} Rows use the same ImageNet validation example and trajectory index from the latent-analysis dataset, with fixed noise fractions $\rho=0.1,0.4,0.8$. Columns show the sequential U-turn index $n$. The qualitative drift accelerates as the per-step noise increases, matching the quantitative collapse of latent persistence.}
%     \label{fig:app_image_sequential_montage}
% \end{figure}

\subsection{Layer-wise persistence across noise levels}

For the quantitative image experiments, we evaluate $20$ ImageNet validation images and noise steps $t\in\{0,100,200,400,600,800,999\}$. For nonzero noise, trajectories have length $N=100$; the $t=100,\ldots,999$ levels use $20$ trajectories per image. ConvNeXt activations are whitened layer-wise using fixed empirical statistics, flattened, and compared to the initial image by cosine similarity. Curves are first averaged within each image and then across images. As in the main image figure, we exclude the classifier head and order layers by ConvNeXt feature depth.

\begin{figure}[htbp]
    \centering
    \includegraphics[width=\linewidth]{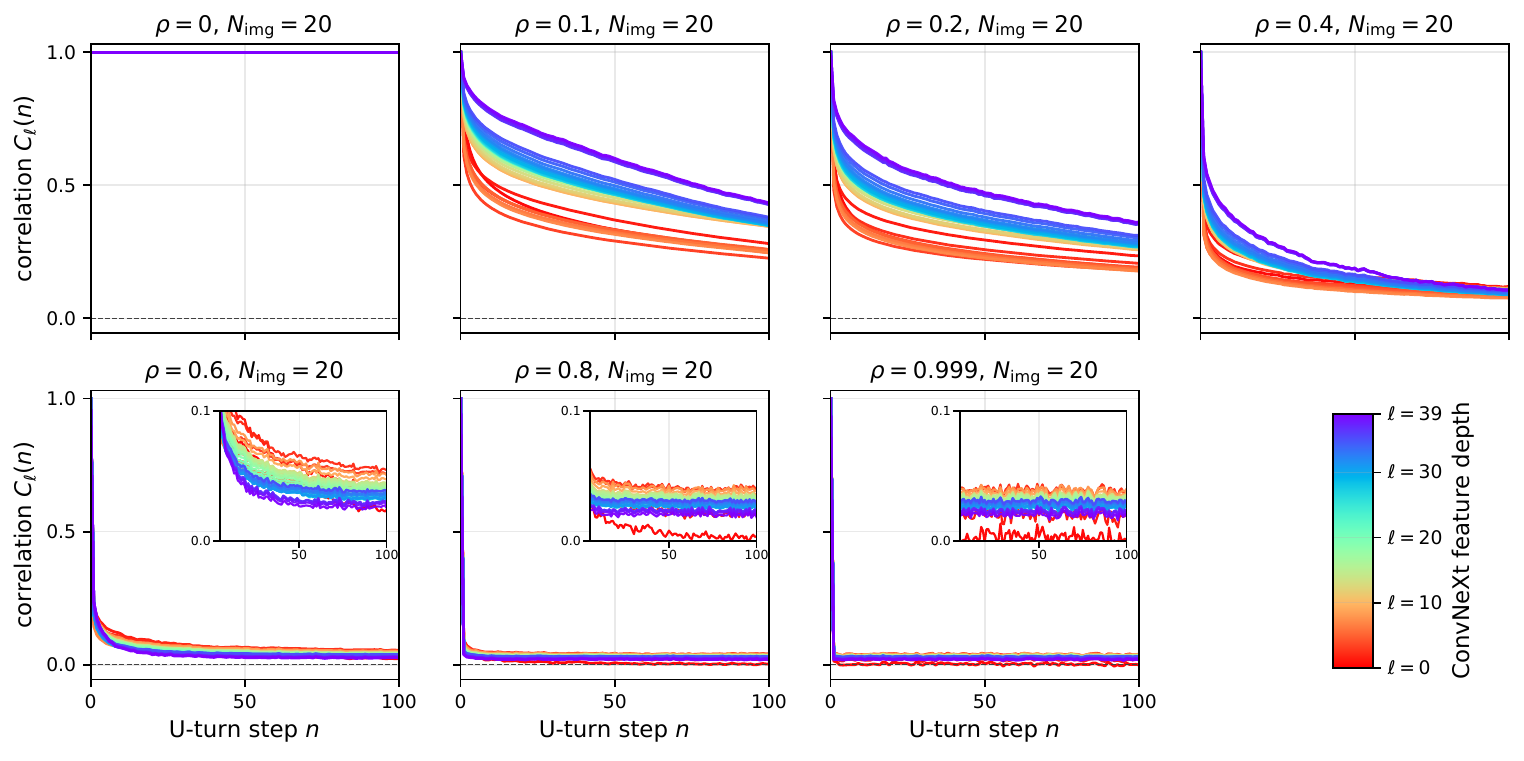}
    \caption{\textbf{Full layer-wise latent persistence sweep.} Cosine correlation between ConvNeXt feature activations at U-turn step $n$ and at initialization, averaged over the 20-image ImageNet validation set with the classifier head excluded. Panels show different noise fractions $\rho=t/T$; insets zoom into the low-correlation regime for large $\rho$. The sweep shows the gradual collapse of long-lived high-level memory as the U-turn magnitude increases.}
    \label{fig:app_image_latent_noise_grid}
\end{figure}

\Cref{fig:app_image_latent_noise_grid} shows the full noise sweep using the same colorbar convention as the main text. Small $\rho$ produces separated relaxation curves, with later visual layers preserving memory longer. As $\rho$ increases, all layers decorrelate more rapidly and the layer ordering narrows. The insets zoom into $0\leq C_\ell(n)\leq0.1$ for the large-noise panels, where the curves are close together but still show the ordering change.

\subsection{Integrated persistence and single-step response}

To summarize the full relaxation curves, we compute the area under each cosine-survival curve on the normalized U-turn axis $s=n/N\in[0,1]$. Let $z_{\ell}^{(i,r)}(n;\rho)$ denote the whitened ConvNeXt feature vector at feature depth $\ell$ after $n$ U-turns for image $i$, trajectory $r$, and noise fraction $\rho$. The per-image, per-layer correlation curve is
\[
C_{\ell}^{(i)}(n;\rho)
=
\frac{1}{R_{i,\rho}}
\sum_{r=1}^{R_{i,\rho}}
\frac{\left\langle z_{\ell}^{(i,r)}(n;\rho), z_{\ell}^{(i)}(0)\right\rangle}
{\left\|z_{\ell}^{(i,r)}(n;\rho)\right\|\,\left\|z_{\ell}^{(i)}(0)\right\|},
\]
where $R_{i,\rho}$ is the number of trajectories for that image and noise level. We include the initial condition as $C_{\ell}^{(i)}(0;\rho)=1$ and define the layer AUC by
\[
A_{\ell}^{(i)}(\rho)
=
\int_{0}^{1} C_{\ell}^{(i)}(sN;\rho)\,ds,
\]
implemented by linear interpolation of the measured curve to a common grid in $s$ followed by trapezoidal integration.

\begin{figure}[htbp]
    \centering
    \includegraphics[width=0.82\linewidth]{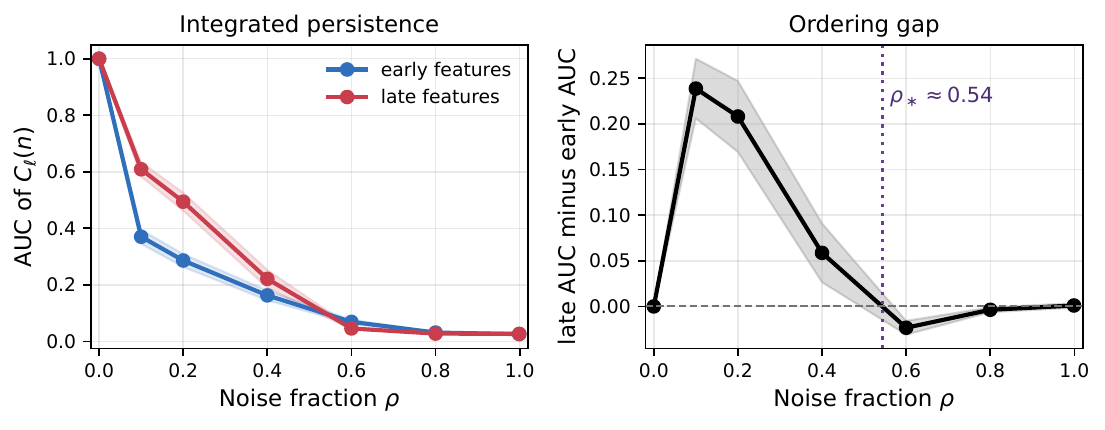}
    \caption{\textbf{AUC summary of image latent persistence.} \textbf{Left:} Mean area under the cosine-survival curve for early and late ConvNeXt feature groups as a function of noise fraction. Shaded bands denote SEM across the 20 images. \textbf{Right:} Difference between late-feature and early-feature AUC. The dotted line marks the interpolated ordering transition for this integrated diagnostic.}
    \label{fig:app_image_auc}
\end{figure}

We sort ConvNeXt outputs by feature depth and exclude the classifier head for the main AUC summary, leaving $L=40$ feature layers indexed $\ell=0,\ldots,L-1$. The early group is the first three features,
\[
\mathcal{G}_{\mathrm{early}}=\{0,1,2\},
\]
and the late group is the last three features,
\[
\mathcal{G}_{\mathrm{late}}=\{L-3,L-2,L-1\}=\{37,38,39\}.
\]
For a group $\mathcal{G}$ we average the layer AUCs,
\[
A_{\mathcal{G}}^{(i)}(\rho)
=
\frac{1}{|\mathcal{G}|}
\sum_{\ell\in\mathcal{G}} A_{\ell}^{(i)}(\rho),
\]
then report the mean and SEM of $A_{\mathcal{G}}^{(i)}(\rho)$ across the $20$ images. The ordering diagnostic in the right panel is
\[
\Delta_{\mathrm{AUC}}(\rho)
=
\left\langle A_{\mathcal{G}_{\mathrm{late}}}^{(i)}(\rho)
- A_{\mathcal{G}_{\mathrm{early}}}^{(i)}(\rho)\right\rangle_i .
\]
Thus $\Delta_{\mathrm{AUC}}>0$ means late, higher-level features retain memory longer than early features, while $\Delta_{\mathrm{AUC}}<0$ indicates the inverted ordering. The dotted vertical line marks the linearly interpolated zero crossing of this gap. This phenomenology is clearly observed in~\cref{fig:app_image_auc}.

\begin{figure}[htbp]
    \centering
    \includegraphics[width=0.68\linewidth]{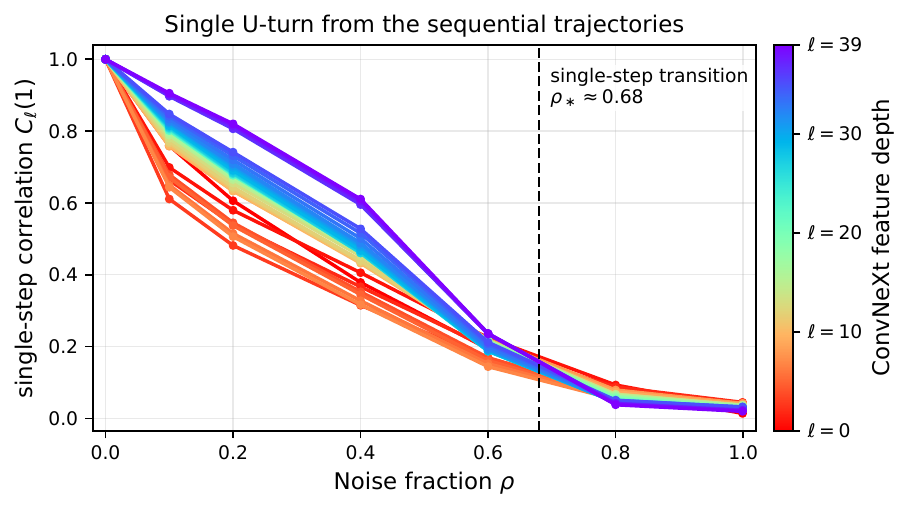}
    \caption{\textbf{Single-U-turn latent response as a function of noise.} Cosine similarity after the first U-turn, $C_\ell(1)$, plotted for each recorded ConvNeXt feature layer using the sequential trajectory dataset. The classifier head is excluded. The dashed line marks the single-step layer-ordering transition.}
    \label{fig:app_image_single_uturn_latents}
\end{figure}

\Cref{fig:app_image_single_uturn_latents} gives a complementary single-step diagnostic by extracting $C_\ell(1)$ from the first step of the same sequential U-turn trajectories used above. This view makes clear that increasing $\rho$ first perturbs early layers while leaving deeper representations relatively stable, and then eventually affects all recorded feature layers. The dashed line marks the zero crossing of the late-minus-early single-step gap.

\subsection{Classifier-head and averaging checks}

\begin{figure}[htbp!]
    \centering
    \includegraphics[width=\linewidth]{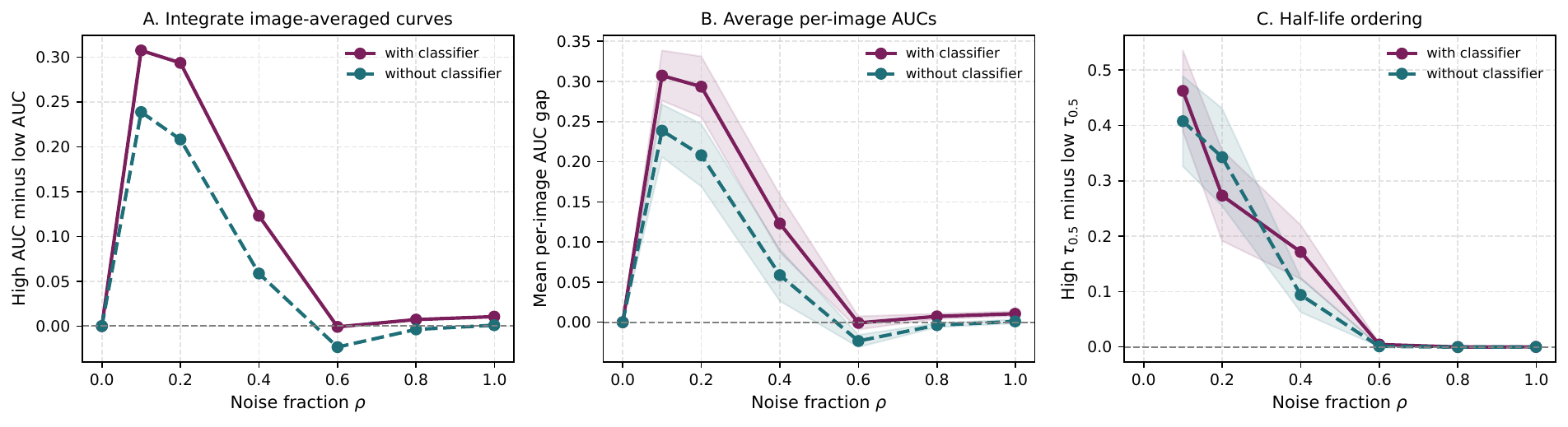}
    \caption{\textbf{Robustness of the image layer-ordering diagnostic.} Ordering summaries are shown both when the classifier head is treated as the deepest output and when it is excluded. The sign and scale of the AUC and half-life gaps show that the observed transition is not an artifact of a single averaging convention or of the classifier head alone.}
    \label{fig:app_image_ordering_variants}
\end{figure}

The main image persistence figure excludes the classifier head so that the ordering is measured only within the ConvNeXt feature hierarchy. As a check, \Cref{fig:app_image_ordering_variants} compares ordering summaries with and without the classifier head and with two equivalent averaging conventions. The qualitative transition is unchanged: high-level memory dominates at small $\rho$, while the ordering weakens and can invert at larger $\rho$.

\subsection{Implementation details}

All image U-turn experiments use the \texttt{guided-diffusion} ImageNet model at $256\times256$ resolution. Input images are resized to the diffusion resolution and scaled to $[-1,1]$. For a U-turn at noise step $t$, the forward step uses the model's standard Gaussian noising process $q(x_t\mid x_0)$; the backward step uses the same ancestral denoising kernel as unconditional sampling, initialized at the forward-noised state rather than at pure Gaussian noise.

For sequential trajectories, the output image of one U-turn becomes the input image to the next U-turn. We save activations at every U-turn step and include the initial state as $C_\ell(0)=1$. The zero-noise point in the sweep is treated analytically as the identity map. AUCs are computed after linear interpolation of each curve onto a common normalized time grid, and SEMs are computed across images after per-image averaging.

% \newpage 

\section{Theory for minimal U-turns in the RHM}
\label{appendix: rhm theory}
\subsection{Ergodic baseline of correlation plateau}
\label{app: rhm setup and observable}

We denote the latent variables at layer \(\ell\) by a vector \(\vec{x}_{\,\ell}\in \mathbb{R}^{s^{L-\ell}}\). Initially, each component satisfies \(x^i_{\,\ell}(t{=}0)\in\{1,\dots,v\}\) for \(i\in\{1,\dots,s^{L-\ell\}}\). The leaf layer is \(\vec{x}_{0}\in\mathbb{R}^{s^L}\). Subscripts denote layers and superscripts denote positions within a layer.

Our main observable is the two-point correlation
\[
      C_{\ell}(n)=\overline{\frac{1}{s^{L-\ell}}\sum_i\Big\langle \delta_{x^i_{\,\ell}(0),\,x^i_{\,\ell}(n)}\Big\rangle},
\]
where \(\delta\) is the Kronecker delta. Here \(\langle \cdot \rangle\) denotes an average over trajectories for a fixed rule, while \(\overline{(\cdot)}\) denotes an average over rules.

% From \(C_{\ell}(n)\), we extract the plateau value \(C_{\ell}(\infty)\) and the relaxation time \(\tau_{\ell}\), defined as the number of time steps required to reach the plateau. 

% \noindent \textbf{Ergodic baseline of the plateau.}
In the ergodic limit, the plateau equals the mean overlap between two independently drawn configurations admissible under a fixed rule. While one might expect \(1/v\), hierarchical constraints induce correlations that increase this value. The correction is controlled by \(f=m/v^{s-1}\), with smaller \(f\) yielding larger overlap.

The ergodic mean overlap is
\[
\begin{aligned}
\overline{\mu_{\ell}}
&=
\frac{1}{v}\left[\frac{r}{1-a} + \left(1-\frac{r}{1-a}\right)a^{L-\ell}\right]
+
\left(1-\frac{1}{v}\right)\frac{r}{1-a}(1-a^{L-\ell}) \\
&\text{where}\quad
r:=\frac{v^{s-1}-1}{v^s-1}, \qquad
a:=\frac{1}{m}(1-r).
\end{aligned}
\]

We define
\[
\tilde C_\ell(n) := \frac{C_\ell(n)-\overline{\mu_\ell}}{1-\overline{\mu_\ell}}.
\]

% \noindent \textbf{Fixed-rule standard deviation of the ergodic sample mean overlap.}
% The expression above gives the disorder-averaged ergodic mean overlap. In simulations,
% however, the plateau is estimated from a finite number \(T\) of independently sampled
% pairs for a fixed realization of the rules. We therefore also estimate the fixed-rule,
% or within-rule, standard deviation of this empirical mean.

% Under the
% approximation that the match indicators at different positions are independent
% Bernoulli variables with success probability \(\overline{\mu_{\ell}}\), we obtain
% \[
%       \mathrm{Var}\!\left(\widehat{\mu}_{\ell}\right)
%       \approx
%       \frac{\overline{\mu_{\ell}}\left(1-\overline{\mu_{\ell}}\right)}
%            {T\,s^{L-\ell}} .
% \]
% Equivalently,
% \[
%       \mathrm{Std}\!\left(\widehat{\mu}_{\ell}\right)
%       \approx
%       \sqrt{
%       \frac{\overline{\mu_{\ell}}\left(1-\overline{\mu_{\ell}}\right)}
%            {T\,s^{L-\ell}}
%       } .
% \]

% For the leaf overlap, \(\ell=0\), this reduces to
% \[
%       \mathrm{Std}\!\left(\widehat{\mu}_{0}\right)
%       \approx
%       \sqrt{
%       \frac{\overline{\mu_{0}}\left(1-\overline{\mu_{0}}\right)}
%            {T\,s^{L}}
%       } .
% \]
\subsection{Finite-size threshold for minimal U-turn}
\label{app:finite threshold}

Here we give the finite-\(v\), finite-\(L\) version of the branching estimate used
to draw the theoretical thresholds in Fig.~\ref{fig:rhm_ergodicity} and
Fig.~\ref{fig:phase_diagram_ergodicity}. The logic is the same as in Sec.~3.5:
starting from one accepted leaf flip, we estimate how many additional leaf flips
it opens. The finite-size calculation keeps two ingredients explicit that are
combined into powers of \(f\) in the main text: the number of admissible
non-original reconstructions, and the probability that such a reconstruction
cascades upward through the hierarchy.

\paragraph{Finite-size admissible values and cascade probabilities.}
Consider a minimal U-turn step, in which a single leaf token is masked and
reconstructed. For each layer \(\ell\), let \(\overline{N_\ell}\) denote the
average number of admissible values of a given layer-\(\ell\) variable, including
its original value, conditioned on fixing all other variables at the same layer
\(\ell\) to their values in the original sample.

The root has no same-layer constraints, so \(\overline{N_L}=v\). Moving downward
from a parent layer to its children gives the recursion derived in
Ref.~\cite{cagnetta2024towards}:
\begin{equation}
\boxed{
\begin{aligned}
    \overline{N_L} &= v, \\
    \overline{N_{\ell-1}}
    &= 1+(v-1)\frac{m\overline{N_\ell}-1}{v^s-1},
    \qquad \ell=1,\ldots,L .
\end{aligned}}
\label{eq:N_recursion_box}
\end{equation}
Thus, at the visible layer, a masked leaf has on average
\(\overline{N_0}-1\) non-original admissible reconstructions, conditioned on all
other visible tokens being fixed.

These admissible-value counts also determine the probability that a change
cascades upward. If a variable at layer \(\ell-1\) changes, its parent at layer
\(\ell\) remains unchanged only when the new configuration is still compatible
with the original parent value. Averaging over admissible parent values, this
happens with probability \(1/\overline{N_\ell}\). Therefore the probability that
the change propagates from layer \(\ell-1\) to layer \(\ell\) is
\begin{equation}
    p^{\mathrm{cascade}}_{\ell-1\to \ell}
    =
    1-\frac{1}{\overline{N_\ell}} .
\label{eq:one_step_cascade}
\end{equation}
The probability that an admissible leaf flip propagates up to level \(\ell\) is
then
\begin{equation}
    p^{\mathrm{cascade}}_{0\to \ell}
    =
    \prod_{r=1}^{\ell}
    \left(1-\frac{1}{\overline{N_r}}\right).
\label{eq:cascade_factor}
\end{equation}

\paragraph{Relation to the main text, Sec.~\ref{sec:branching_estimates}.}
We now spell out how these finite-size quantities correspond to the factors used
in Sec.~\ref{sec:branching_estimates}. As in the main text, the first flip is
already conditioned on being admissible. Under this conditioning, the probability
that it propagates to at least level \(\ell-1\) is
\begin{equation}
    \frac{f^\ell+f^{\ell+1}+\cdots}
         {f+f^2+\cdots}
    =
    f^{\ell-1}.
\label{eq:main_text_first_flip_factor}
\end{equation}
In the finite-size expression, the corresponding conditional probability is
\(p^{\mathrm{cascade}}_{0\to \ell-1}\).

In the main text, the second-flip factor
\begin{equation}
    f^\ell+f^{\ell+1}+\cdots
    =
    \frac{f^\ell}{1-f}
\label{eq:main_text_second_flip_factor}
\end{equation}
combines two pieces: the number of non-original admissible reconstructions and
the probability that such a reconstruction propagates to at least level
\(\ell-1\). In the finite-size expression, we keep these two pieces separate as
\begin{equation}
    \left(\overline{N_0}-1\right)
    p^{\mathrm{cascade}}_{0\to \ell-1}.
\label{eq:finite_second_flip_factor}
\end{equation}

\paragraph{Mean number of novel flips in Finite-size.}
Fix an accepted leaf flip at position \(i\), and consider another leaf \(j\)
whose lowest common ancestor with \(i\) is at layer \(\ell\). For fixed \(i\),
the number of such leaves is \(s^\ell-s^{\ell-1}\).

For the flip at \(i\) to open a move at \(j\), the first flip must affect the
child of the common ancestor that lies on the path from \(i\). Equivalently, its
cascade must reach level \(\ell-1\). This gives the finite-size version of the
first main-text factor:
\begin{equation}
    p^{\mathrm{cascade}}_{0\to \ell-1}.
\end{equation}

After the first flip, stationarity implies that leaf \(j\) has on average
\(\overline{N_0}-1\) non-original admissible alternatives. However, only those
alternatives whose cascade also reaches level \(\ell-1\) can couple to the
change created by the first flip. This gives the finite-size version of the
second main-text factor:
\begin{equation}
    \left(\overline{N_0}-1\right)
    p^{\mathrm{cascade}}_{0\to \ell-1}.
\end{equation}
Together, the two cascade requirements contribute
\begin{equation}
    \left(\overline{N_0}-1\right)
    \left(p^{\mathrm{cascade}}_{0\to \ell-1}\right)^2 .
\end{equation}

There is one additional finite-size correction. Even when both cascades reach
level \(\ell-1\), the candidate value at that level may already have been
admissible regardless of the first flip. To count only newly opened moves, we
approximate the admissible alternative set at layer \(\ell-1\) as a uniformly
random subset of size \(\overline{N_{\ell-1}}-1\) among the \(v-1\) non-original
values. Under this approximation, the probability that the candidate value was
already admissible is
\begin{equation}
    \frac{\overline{N_{\ell-1}}-1}{v-1},
\end{equation}
so the probability that it is newly enabled by the first flip is
\begin{equation}
    1-
    \frac{\overline{N_{\ell-1}}-1}{v-1}.
\label{eq:newly_enabled_probability}
\end{equation}

Summing over all possible positions \(j\), we obtain the finite-\(v\), finite-\(L\)
mean number of newly opened flips:
\begin{equation}
\boxed{
\begin{aligned}
n_{v,L}(f)
:=
\sum_{\ell=1}^{L}
\left(s^\ell-s^{\ell-1}\right)
\left(\overline{N_0}-1\right)
\left(p^{\mathrm{cascade}}_{0\to \ell-1}\right)^2
\left[
    1-
    \frac{\overline{N_{\ell-1}}-1}{v-1}
\right] .
\end{aligned}}
\label{eq:n_finite}
\end{equation}

\paragraph{Percolation threshold.}
As in the main text, the percolation threshold is estimated by the condition that
one accepted flip opens, on average, one additional admissible flip:
\begin{equation}
    n_{v,L}\!\left(f_{\mathrm{per}}^{(v,L)}\right)=1 .
\label{eq:fper_finite}
\end{equation}
For Fig.~\ref{fig:rhm_ergodicity} and Fig.~\ref{fig:phase_diagram_ergodicity},
we use \eqref{eq:n_finite}: we first compute the
\(\overline{N_\ell}\) recursively from \eqref{eq:N_recursion_box}, then
evaluate the cascade probabilities using \eqref{eq:cascade_factor}, and
finally solve \eqref{eq:fper_finite}.

\paragraph{Layer-ordering inversion threshold.}
The inversion estimate uses the same finite-size cascade probability. The
ordering between adjacent layers is controlled by two competing effects. First,
layer \(\ell\) contains a factor \(s\) fewer variables than layer \(\ell-1\),
which favors faster decorrelation at higher layers. Second, a change at layer
\(\ell-1\) reaches layer \(\ell\) only with probability
\(p^{\mathrm{cascade}}_{\ell-1\to\ell}\), which favors slower decorrelation at
higher layers. Balancing these two effects gives
\begin{equation}
    s\,p^{\mathrm{cascade}}_{\ell-1\to\ell} = 1,
    \qquad
    p^{\mathrm{cascade}}_{\ell-1\to\ell}
    =
    1-\frac{1}{\overline{N_\ell}} .
\label{eq:finv_general}
\end{equation}
In Fig.~\ref{fig:phase_diagram_ergodicity}, we compare the visible layer and the
first latent layer, so we use
\begin{equation}
    s\,p^{\mathrm{cascade}}_{0\to 1}
    =
    s\left(1-\frac{1}{\overline{N_1}}\right)
    =1
\label{eq:finv_finite}
\end{equation}
to compute the finite-\(v\), finite-\(L\) threshold
\(f_{\mathrm{inv}}^{(v,L)}\).

\paragraph{Asymptotic limit.}
We now take the limit \(v\to\infty\) and \(L\to\infty\) at fixed rule density
\(f=m/v^{s-1}\). In this
limit,
\begin{equation}
    \overline{N_\ell}
    \longrightarrow
    \frac{1}{1-f},
\label{eq:N_asymptotic}
\end{equation}
so that the one-step cascade probability becomes
\begin{equation}
    p^{\mathrm{cascade}}_{\ell-1\to \ell}
    =
    1-\frac{1}{\overline{N_\ell}}
    \longrightarrow
    f .
\label{eq:one_step_cascade_asymptotic}
\end{equation}
Consequently,
\begin{equation}
    p^{\mathrm{cascade}}_{0\to \ell}
    =
    \prod_{r=1}^{\ell}
    \left(1-\frac{1}{\overline{N_r}}\right)
    \longrightarrow
    f^\ell .
\label{eq:cascade_asymptotic}
\end{equation}
Applying these limits to \eqref{eq:n_finite} gives
\begin{equation}
    n_{v,L}(f)
    \longrightarrow
    \sum_{\ell=1}^{\infty}
    \left(s^\ell-s^{\ell-1}\right)
    \frac{f}{1-f}
    \left(f^{\ell-1}\right)^2
    =
    \frac{f(s-1)}{(1-sf^2)(1-f)} .
\label{eq:n_asymptotic}
\end{equation}
The percolation condition then reduces to the main-text estimate
\begin{equation}
    f_{\mathrm{per}}
    =
    \frac{1}{s}
    -
    \frac{1}{s^2}
    +
    O\!\left(\frac{1}{s^3}\right),
\label{eq:fper_asymptotic}
\end{equation}
while the inversion condition reduces to the main-text estimate
\begin{equation}
    s f_{\mathrm{inv}} = 1,
    \qquad
    f_{\mathrm{inv}} = \frac{1}{s}.
\label{eq:finv_asymptotic}
\end{equation}

\end{document}